\documentclass[10pt,twocolumn,letterpaper]{article}

\usepackage{cvpr}              
\definecolor{cvprblue}{rgb}{0.21,0.49,0.74}
\usepackage[pagebackref,breaklinks,colorlinks,allcolors=cvprblue]{hyperref}
\usepackage[table]{xcolor}
\usepackage{colortbl} 
\usepackage{array} 
\usepackage{booktabs} 
\usepackage{float}

\newcommand{\xh}{\textcolor[RGB]{0,0,0}}
\def\paperID{8067} 
\def\confName{CVPR}
\def\confYear{2026}
\newcommand{\ours}{{\fontfamily{ppl}\selectfont SurvAgent}}

\title{\raisebox{-0.3\height}{\includegraphics[width=1cm]{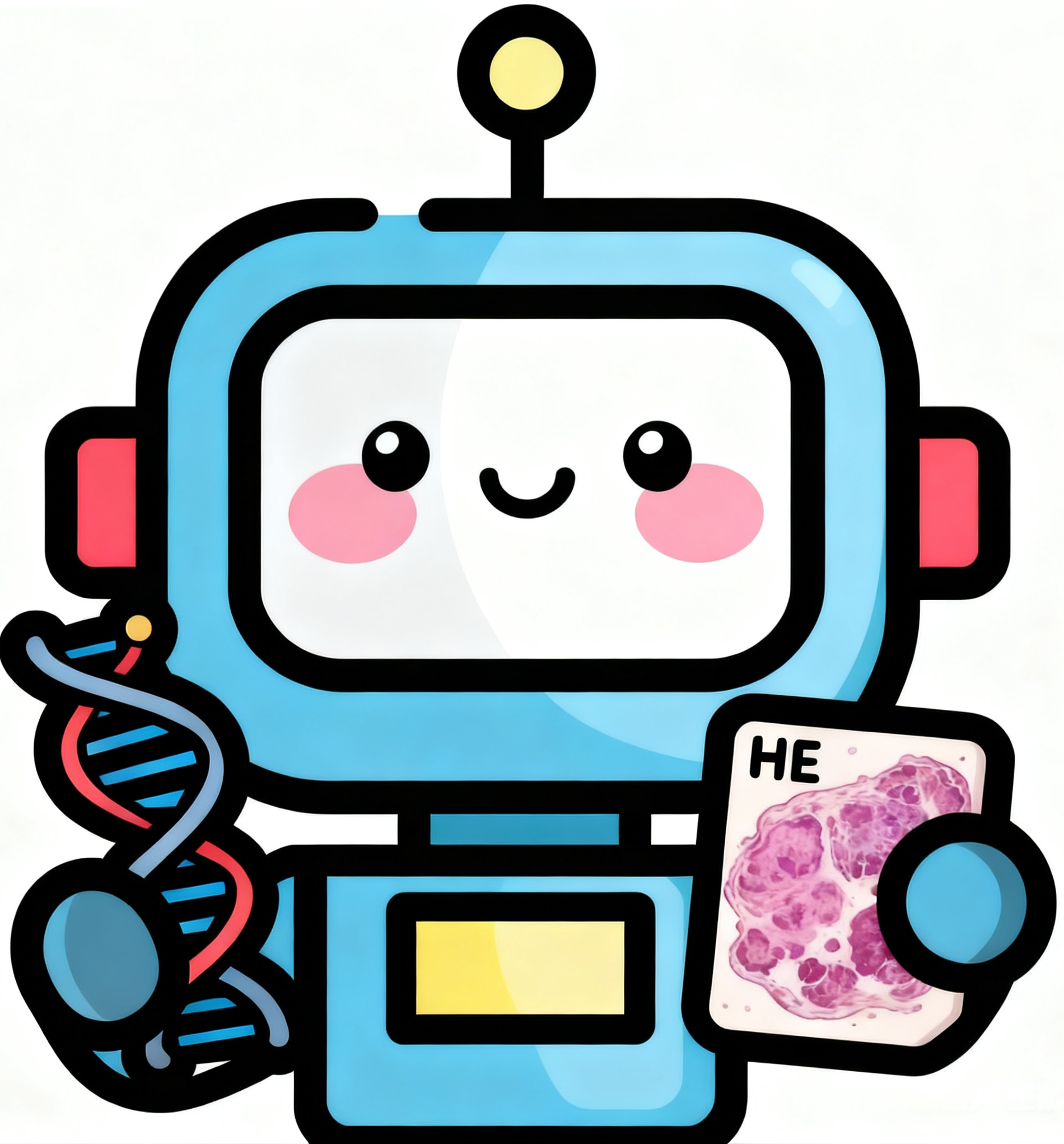}}\ours: Hierarchical CoT-Enhanced Case Banking and\\ Dichotomy-Based Multi-Agent System for Multimodal Survival Prediction}

\author{
    Guolin Huang$^{1}$\thanks {Equal contribution.},
    Wenting Chen$^{2}$\footnotemark[1], 
    Jiaqi Yang$^{3}$, 
    Xinheng Lyu$^{3}$, \\
    Xiaoling Luo$^{1}$,
    Sen Yang$^{4}$, 
    Xiaohan Xing$^{2}$\thanks{Corresponding Author.},
    Linlin Shen$^{1}$\footnotemark[2]\\
    $^{1}$Shenzhen University, 
    $^{2}$Stanford University, 
    $^{3}$University of Nottingham Ningbo China, 
    $^{4}$Ant Group
}

\begin{document}
\maketitle
\begin{abstract}

Survival analysis is critical for cancer prognosis and treatment planning, yet existing methods lack the transparency essential for clinical adoption. While recent pathology agents have demonstrated explainability in diagnostic tasks, they face three limitations for survival prediction: inability to integrate multimodal data, ineffective region-of-interest exploration, and failure to leverage experiential learning from historical cases. We introduce \ours, the first hierarchical chain-of-thought (CoT)-enhanced multi-agent system for multimodal survival prediction. \ours~consists of two stages: (1) WSI-Gene CoT-Enhanced Case Bank Construction employs hierarchical analysis through Low-Magnification Screening, Cross-Modal Similarity-Aware Patch Mining, and Confidence-Aware Patch Mining for pathology images, while Gene-Stratified analysis processes six functional gene categories. Both generate structured reports with CoT reasoning, storing complete analytical processes for experiential learning. (2) Dichotomy-Based Multi-Expert Agent Inference retrieves similar cases via RAG and integrates multimodal reports with expert predictions through progressive interval refinement. Extensive experiments on five TCGA cohorts demonstrate \ours's superority over conventional methods, proprietary MLLMs, and medical agents, establishing a new paradigm for explainable AI-driven survival prediction in precision oncology.

\end{abstract}






\section{Introduction}
\label{sec:intro}

Survival analysis estimates patient survival time using pathology whole slide images (WSIs) and genomic data, providing crucial insights for cancer treatment and precision medicine~\cite{gyHorffy2021survival,kaplan1958nonparametric}. While numerous studies~\cite{jaume2024modeling,zhou2023cross,zhou2024cohort,zhang2024prototypical,song2024morphological,zhou2025robust} have achieved significant performance, existing methods lack transparent decision-making and interpretable explanations, which are essential for clinicians to validate predictions and make informed treatment decisions~\cite{abbas2025explainable,raz2025explainable}. Developing explainable multimodal survival analysis methods is therefore critical for clinical trust and adoption.

\begin{figure}[t]
    \centering
    \includegraphics[width=\linewidth]{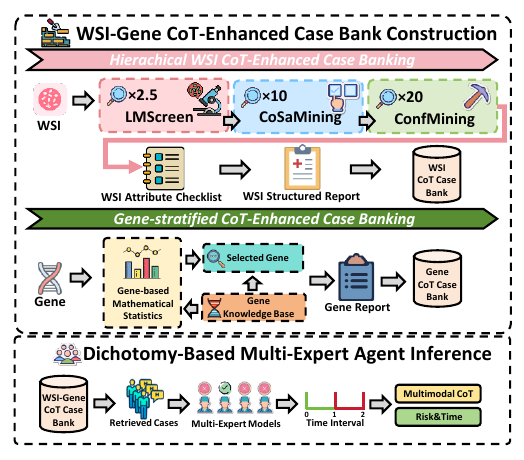}
    \caption{\textbf{\ours}~utilizes WSI-Gene CoT-Enhanced Case Banks through hierarchical WSI analysis and gene-stratified analysis, then performs dichotomy-based multi-expert inference by retrieving cases and progressively refining survival predictions.} 
    \label{fig:km}
\end{figure}


Due to the success of large language model (LLM)-based agents, they are increasingly used in medicine for various scenarios. These agents gather patient evidence, synthesize data, enable cross-specialty collaboration, and provide transparent decision-making. Pathology agents~\cite{quang2025gmat,wang2025pathology,sun2025cpathagent,ghezloo2025pathfinder,chen2025evidence,lyu2025inline} are developed for WSI-based diagnostics, interpretably mimicking pathologists' reasoning through operational actions (e.g., zooming, panning) while explaining their logic. 
However, no agent exists specifically for survival prediction, and adapting diagnostic agents to this task poses major challenges that limit performance.

First, existing pathology agents~\cite{wang2025pathology,sun2025cpathagent,ghezloo2025pathfinder,chen2025evidence,lyu2025inline} primarily accept \textbf{single-modality input} (i.e., WSIs), \xh{which constrains their ability to incorporate genomic data—an essential source of molecular information for understanding tumor biology and improving survival prediction.  
}
\xh{Morphological insights from histopathology and molecular profiles from genomics offer complementary perspectives on tumor evolution and therapeutic response.
Thus, integrating multimodal data within pathology agents is crucial for capturing both phenotypic and molecular determinants of prognosis, ultimately enabling more accurate and biologically informed survival prediction.}

\xh{Another challenge lies in the \textbf{ineffective Region-of-Interest (ROI) exploration strategies} used by current pathology agents.
Expert-based approaches~\citep{wang2025pathology} rely on pathologists’ viewing behaviors to identify ROIs, but this process is labor-intensive and yields limited training data, hindering scalability.
Automatic methods attempt to overcome this issue but introduce new trade-offs: CPathAgent~\citep{sun2025cpathagent} downscales WSIs for efficiency at the cost of fine-grained lesion details, PathFinder~\citep{ghezloo2025pathfinder} preserves high resolution but requires time-consuming sequential patch selection, and SlideSeek~\citep{chen2025evidence} employs tissue detection yet often captures irrelevant or incomplete tumor regions.
Overall, existing methods either miss critical lesions, demand excessive computation, or include redundant areas, highlighting the need for an ROI mining strategy that balances accuracy, efficiency, and coverage for robust survival prediction.}

\xh{Third, current pathology agents primarily analyze each test case in isolation, \textbf{overlooking valuable prognostic experience} from previous cases.
Most existing agents~\citep{wang2025pathology,sun2025cpathagent,ghezloo2025pathfinder,chen2025evidence,lyu2025inline} rely on pathology foundation models or case-specific knowledge bases to process WSIs independently, without leveraging information from similar patients.
In contrast, clinicians often estimate survival by referencing comparable cases, highlighting the need to incorporate experiential knowledge.
Although some general-purpose medical agents~\citep{li2025care,li2024agent,jiang2024long,liu2024medchain} introduce memory or database mechanisms, they mainly store factual data while neglecting the reasoning process—how clinicians integrate and weigh evidence to reach diagnostic conclusions~\citep{donroe2024clinical,rutter2025importance,choi2023using}.
Thus, current agents need to reconstruct reasoning patterns for each new case, underscoring the importance of integrating experiential learning with explicit reasoning pathways for effective survival prediction.}

\xh{To address these challenges, we introduce \textbf{\ours}, the first multi-agent system specifically designed for multimodal survival prediction. Our framework consists of two key components: (1) \textbf{WSI-Gene CoT-Enhanced Case Bank Construction} for experiential learning, and (2) \textbf{Dichotomy-Based Multi-Expert Agent Inference} for final prediction.}

\textbf{WSI-Gene CoT-Enhanced Case Bank Construction} generates reasoning-based case analyses with explicit pathways for WSI and genomic data via two modules. For \textit{\textbf{effective WSI ROI exploration}}, the \textit{Hierarchical WSI CoT-Enhanced Case Bank} uses a multi-magnification pipeline: (1) At low magnification, \textit{Low-Magnification Screening (LMScreen)}—PathAgent generates global WSI reports; (2) At medium magnification, \textit{Cross-Modal Similarity-Aware Patch Mining (CoSMining)}—excludes redundant patches by computing self-patch and self-report similarity, selecting patches meeting both criteria; (3) At high magnification, \textit{Confidence-Aware Patch Mining (ConfMining)}—identifies low-confidence patches, zooms in, and applies CoSMining to capture overlooked lesions. PathAgent extracts attributes via a pre-defined WSI attribute checklist, generates structured reports, and creates CoT reasoning from reports and ground-truth survival times. To \textit{\textbf{incorporate experiential learning with explicit reasoning pathways}}, a self-critique and refinement mechanism stores not only patient facts but also the complete analytical reasoning process for survival prediction. The WSI CoT Case Bank stores CoTs, summarized reports, and survival times, enabling future cases to benefit from both diagnostic conclusions and reasoning processes of similar historical cases. For \textbf{\textit{multimodal data integration}}, the \textit{Gene-Stratified CoT-Enhanced Case Bank} has GenAgent analyze genomics by classifying genes into six types, performing statistical analysis, and generating type-specific reports using a knowledge base. After summarization, CoT generation, and self-critique, the Gene CoT Case Bank stores CoTs, summarized reports, and survival times alongside pathological data.

\textbf{Dichotomy-Based Multi-Expert Agent Inference} stage leverages the constructed case banks for final prediction. For test cases, the system generates hierarchical WSI reports through LMScreen, CoSMining, and ConfMining, while genomic reports are produced using the same pipeline as during case bank construction. Using retrieval-augmented generation (RAG), similar cases with their stored reasoning pathways are retrieved based on multimodal report similarity, allowing the system to reference both the conclusions and analytical processes from comparable historical cases. An inference agent then integrates retrieved cases, summarized reports, and predictions from multiple expert survival models. Rather than directly predicting survival time, the agent employs \textit{dichotomy-based reasoning}: first classifying the case into coarse survival intervals, then progressively refining the classification, and finally predicting exact survival time within the identified interval. This hierarchical decision process, combined with comprehensive WSI-gene reports and explicit reasoning pathways, provides transparent and interpretable survival predictions that align with clinical decision-making practices. Our contributions are summarized as follows:
\begin{itemize}
    \item We propose \ours, the first multi-agent system for multimodal survival prediction, featuring WSI-Gene CoT-Enhanced Case Banking that stores analytical reasoning processes to enable experiential learning.

    
    \item We introduce Hierarchical WSI CoT-Enhanced Case Bank with LMScreen, CoSMining, and ConfMining that balances accuracy, efficiency, and coverage across multiple magnifications.

    \item We propose Dichotomy-Based Multi-Expert Agent Inference that integrates retrieved cases, multimodal reports, and expert predictions through progressive interval refinement for transparent survival time prediction.

    \item Extensive experiments demonstrate that SurvAgent achieves the best C-index across 5 datasets while providing interpretable multimodal reports and CoT reasoning.

\end{itemize}


\begin{figure*}[!t]
\centering
\includegraphics[width=0.95\textwidth]{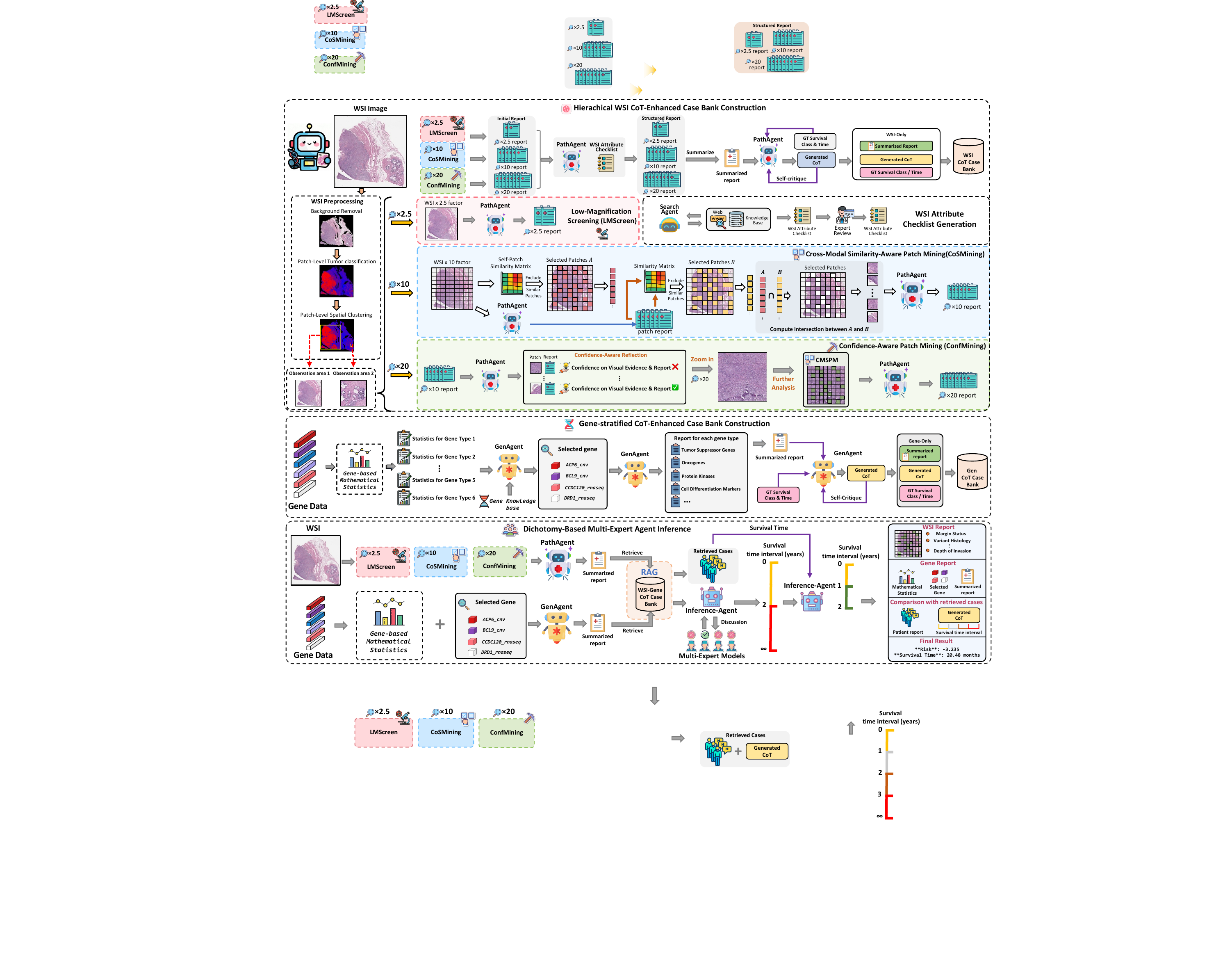}
\caption{Overview of \ours. (1) \textbf{WSI-Gene CoT-Enhanced Case Bank Construction} includes \textit{Hierarchical WSI CoT-Enhanced Case Bank} that progressively analyzes WSIs at multiple magnifications through LMScreen, CoSMining, and ConfMining, and \textit{Gene-Stratified CoT-Enhanced Case Bank} for gene statistical analysis. PathAgent and GenAgent generate structured reports and CoT reasoning with self-critique for their respective case banks. (2) \textbf{Dichotomy-Based Multi-Expert Agent Inference} uses RAG for retrieval and integrates retrieved cases, reports, and expert predictions for progressive survival time prediction from coarse to fine-grained intervals.}
\label{overveiw}
\end{figure*}



\section{Related Works}

%
\subsection{Multimodal Survival Prediction}

Current multimodal survival analysis approaches~\cite{jaume2024modeling,zhou2023cross,zhou2024cohort,zhang2024prototypical,song2024morphological,zhou2025robust,chen2020pathomic,chen2021multimodal,xu2023multimodal,zhang2024prototypical} integrate pathological WSIs and genomic data to provide a more comprehensive perspective on patient stratification and prognosis. For instance, Chen et al.\cite{chen2020pathomic} employ the Kronecker product to model pairwise feature interactions across multimodal data. Chen et al.\cite{chen2021multimodal} present a multimodal co-attention transformer (MCAT) framework that learns an interpretable, dense co-attention mapping between WSIs and genomic features within a unified embedding space. 
\xh{However, they achieve strong multimodal survival prediction performance at the expense of interpretability, lacking the transparent reasoning required for clinical trust and adoption~\cite{abbas2025explainable,raz2025explainable}.
Thus, we propose a multi-agent system that enables reliable and explainable survival analysis through transparent decision-making.}

\subsection{Medical LLM-based Agent}

LLM-based agents in medicine include general-purpose systems and specialty-designed agents for radiology~\cite{lou2025cxragent, tzanis2025agentic}, 
gastroenterology~\cite{tang2025endoagent}, 
oncology~\cite{ferber2025development}, and pathology~\cite{quang2025gmat,wang2025pathology,sun2025cpathagent,ghezloo2025pathfinder,chen2025evidence,lyu2025inline}. The pathology agents perform operational actions while articulating analytical logic, but face limitations for survival prediction: they process only WSIs without integrating genomic data~\cite{wang2025pathology,sun2025cpathagent,ghezloo2025pathfinder,chen2025evidence,lyu2025inline}, their ROI exploration involves trade-offs between labor-intensive expert annotation~\cite{wang2025pathology} and automatic approaches with reduced resolution~\cite{sun2025cpathagent}, excessive time~\cite{ghezloo2025pathfinder}, or irrelevant regions~\cite{chen2025evidence}, and they analyze cases in isolation. While general-purpose agents~\cite{li2025care,li2024agent,jiang2024long,liu2024medchain} use memory mechanisms, they store only factual data without reasoning processes. Thus, we propose a dichotomy-based multi-agent framework with cross-modal analysis, efficient patch mining for ROI selection, and WSI-Gene CoT-enhanced case banking to preserve complete prognostic reasoning.
\section{Method}
\label{sec:method}

In Fig.~\ref{overveiw}, \ours~comprises two stages: WSI-Gene CoT-Enhanced Case Bank Construction and Dichotomy-Based Multi-Expert Agent Inference.
In the first stage, we build case banks with pathological and genomic reasoning pathways. For WSI analysis, a hierarchical pipeline operates at increasing magnifications: LMScreen generates global reports at $\times 2.5$, CoSMining excludes redundant patches while retaining important regions at $\times 10$ via self-patch and self-report similarity, and ConfMining identifies low-confidence patches at $\times 20$ for further exploration. PathAgent extracts attributes using a pre-defined checklist, generates structured reports, and creates CoT reasoning with self-critique. CoTs, reports, and survival times are stored in the WSI CoT Case Bank. For genomic data, GenAgent performs gene-stratified analysis and generates type-specific reports, stored in the Gene CoT Case Bank.
During inference, test cases undergo hierarchical analysis to generate multimodal reports. Using RAG, similar cases with stored CoTs are retrieved. Inference agent integrates retrieved cases, multimodal reports, and expert predictions through dichotomy-based reasoning, progressively refining survival intervals to predict exact survival time.

\subsection{WSI-Gene CoT-Enhanced Case Bank Construction }
\xh{To enable experiential learning with explicit reasoning, we construct a WSI-Gene CoT-Enhanced Case Bank that stores both patient facts and reasoning traces, mimicking how clinicians draw on experience from similar cases.}
It comprises two components: (1) Hierarchical WSI CoT-Enhanced Case Bank performs multi-magnification analysis through LMScreen, CoSMining, and ConfMining to generate pathological reports and reasoning pathways, and (2) Gene-Stratified CoT-Enhanced Case Bank conducts systematic analysis across six gene types to produce genomic reports and reasoning pathways. Both components generate summarized reports, create CoT reasoning based on ground-truth survival times, apply self-critique for refinement, and store triplets into case banks for retrieval during inference.

\subsubsection{Hierachical WSI CoT-Enhanced Case Bank} 

\noindent\textbf{Low-Magnification Screening (LMScreen).} To obtain comprehensive global understanding of the WSI while maintaining computational efficiency, we perform initial analysis at $\times 2.5$ magnification. Given a WSI $\mathcal{W}$, we first downsample it to obtain the low-magnification representation $\mathcal{W}_{2.5}$. A PathAgent $\mathcal{A}_{\text{wsi}}$ processes this representation to generate a global report:
$\mathcal{R}_{\text{global}} = \mathcal{A}_{\text{wsi}}(\mathcal{W}_{2.5})$. This global report $\mathcal{R}_{\text{global}}$ captures overall tissue architecture. 

\noindent\textbf{Cross-Modal Similarity-Aware Patch Mining (CoSMining).} For effective WSI ROI exploration, we introduce CoSMining to mine fine-grained high-magnification patches by select diagnostically important patches while excluding redundant regions through self-patch and self-report similarities. Specifically, given the downsampled WSI $\mathcal{W}_{10}$ at $\times 10$ magnification, we partition it into $N$ non-overlapping patches $\{\mathbf{p}_i\}_{i=1}^N$. To eliminate visual redundancy, we construct a self-patch similarity matrix $\mathbf{S}^v \in \mathbb{R}^{N \times N}$, where each element is computed as:
$S^v_{ij} = \text{sim}(\phi(\mathbf{p}_i), \phi(\mathbf{p}_j)),$
where $\phi(\cdot)$ denotes a pathology foundation model encoder~\cite{wang2024pathology} and $\text{sim}(\cdot, \cdot)$ represents cosine similarity. We identify and remove patches with high visual similarity by selecting patches whose maximum similarity to others exceeds threshold $\tau_v$:
\xh{$\mathcal{P}^v_{\text{selected}} = \{\mathbf{p}_i \mid \max_{j \neq i} S^v_{ij} < \tau_v\}$.}

To ensure semantic diversity, PathAgent generates preliminary reports $\{\mathcal{R}_i\}_{i=1}^N$ for all patches. We then construct a self-report similarity matrix $\mathbf{S}^t \in \mathbb{R}^{N \times N}$ in the textual semantic space:
$S^t_{ij} = \text{sim}(\psi(\mathcal{R}_i), \psi(\mathcal{R}_j)),$
where $\psi(\cdot)$ represents text embedding encoded by a text encoder~\cite{devlin2019bert}. Similarly, we select semantically diverse patches through the threshold $\tau_t$:
\xh{$\mathcal{P}^t_{\text{selected}} = \{\mathbf{p}_i \mid \max_{j \neq i} S^t_{ij} < \tau_t\}.$}

The final selected patches are obtained through intersection of both criteria:
\xh{$\mathcal{P}_{10} = \mathcal{P}^v_{\text{selected}} \cap \mathcal{P}^t_{\text{selected}},$}
ensuring patches are both visually distinctive and semantically informative. PathAgent then generates detailed reports $\{\mathcal{R}^{10}_k\}_{k=1}^{|\mathcal{P}_{10}|}$ for these selected patches.

\noindent\textbf{Confidence-Aware Patch Mining (ConfMining).} 
After CoSMining identifies potentially informative patches at $10 \times$ magnification, not all patches require further high-magnification analysis. To efficiently allocate computational resources, ConfMining introduces a Confidence-Aware Reflection mechanism that selectively determines which patches warrant deeper examination at $20 \times$ magnification based on PathAgent's analytical confidence. For each patch $\mathbf{p}_k \in \mathcal{P}_{10}$ with its corresponding report $\mathcal{R}^{10}_k$, PathAgent predicts the confidence level of the reports from three categories: low, medium, and high. When PathAgent assigns a low confidence level—indicating uncertainty about morphological features, ambiguous cellular patterns, or the need for finer details—the patch is selected for hierarchical $20 \times$ magnification analysis. 

For each low-confidence patch $\mathbf{p}_k \in \mathcal{P}_{\text{low-conf}}$, we zoom in to $\times 20$ magnification to obtain its high-resolution version $\mathbf{p}^{20}_k$ and partition it into $M$ sub-patches $\{\mathbf{p}^{20}_{k,m}\}_{m=1}^M$. To avoid exhaustive analysis of all sub-patches, we apply the CoSMining strategy at this finer scale to extract the most informative sub-patches through cross-modal similarity filtering:
\xh{$\mathcal{P}^{20}_k = \text{CoSMining}(\mathbf{p}^{20}_k; \tau_v, \tau_t).$}
The final set of high-magnification patches is $\mathcal{P}_{20} = \bigcup_{k} \mathcal{P}^{20}_k$, and PathAgent generates corresponding detailed reports $\{\mathcal{R}^{20}_m\}_{m=1}^{|\mathcal{P}_{20}|}$ for these selected sub-patches. This two-stage confidence-driven and similarity-aware mining process ensures thorough analysis of uncertain regions while maintaining computational efficiency by focusing only on the most relevant fine-grained features within low-confidence patches.

\noindent\textbf{WSI Attribute Checklist Generation.} To ensure structured and clinically relevant analysis, we employ a search agent $\mathcal{A}_{\text{search}}$ to establish a WSI attribute checklist $\mathcal{C}_{\text{WSI}}$. The search agent queries medical knowledge bases $\mathcal{K}_{\text{med}}$~\cite{lyu2025inline} and online resources $\mathcal{D}_{\text{web}}$~\cite{MyPathologyReport_zhTW} to identify prognostically important attributes:
$\mathcal{C}_{\text{raw}} = \mathcal{A}_{\text{search}}(\mathcal{K}_{\text{med}}, \mathcal{D}_{\text{web}}; q_{\text{survival}})$,
where $q_{\text{survival}}$ represents survival-prediction-specific queries. The raw checklist $\mathcal{C}_{\text{raw}}$ is then reviewed and refined by clinical experts to obtain $\mathcal{C}_{\text{WSI}} = \{a_1, a_2, \ldots, a_K\}$, where each $a_k$ represents a key pathological attribute (e.g., tumor grade, necrosis extent, lymphocytic infiltration). We include 16 key attributes.

Using this checklist, PathAgent extracts structured information from all hierarchical reports:
$$\mathcal{R}_{\text{struct}} = \mathcal{A}_{\text{wsi}}(\{\mathcal{R}_{\text{global}}, \{\mathcal{R}^{10}_k\}, \{\mathcal{R}^{20}_m\}\}; \mathcal{C}_{\text{WSI}}),$$
and generates a summarized report $\mathcal{R}_{\text{sum}}^{\text{WSI}}$ by removing redundant information.
Finally, given the summarized report $\mathcal{R}_{\text{sum}}^{\text{WSI}}$ and GT survival time $t_{\text{GT}}$, PathAgent generates CoT reasoning:
$\text{CoT}_{\text{WSI}} = \mathcal{A}_{\text{wsi}}(\mathcal{R}_{\text{sum}}^{\text{WSI}}, t_{\text{GT}}).$

To ensure the correctness of the CoT, we utilize a self-critique mechanism to evaluate the generated CoT. Specifically, we employ Qwen2.5-32B as the quality validation function $\mathcal{V}(\cdot)$ to assess the CoT and generate both a quality level (low or high) and detailed critique. The refinement process is formulated as:
\begin{equation}
\text{CoT}_{\text{WSI}}^{\text{refined}} = \begin{cases}
\text{CoT}_{\text{WSI}} & \text{if } \mathcal{V}(\text{CoT}_{\text{WSI}}) = \text{high} \\
\mathcal{A}_{\text{wsi}}(\text{CoT}_{\text{WSI}}, \text{Critique}) & \text{Otherwise} 
\end{cases},
\end{equation}
where $\mathcal{V}(\text{CoT}_{\text{WSI}})$ returns the quality level and generates a critique, and $\mathcal{A}_{\text{wsi}}(\cdot, \cdot)$ denotes PathAgent's refinement operation that revises the CoT based on critique feedback until high quality is achieved. The final triplet $(\mathcal{R}_{\text{sum}}^{\text{WSI}}, \text{CoT}_{\text{WSI}}^{\text{refined}}, t_{\text{GT}})$ is stored in the WSI CoT Case Bank $\mathcal{B}_{\text{WSI}}$. Through this hierarchical pipeline, from global screening to cross-modal mining and confidence-aware refinement, we construct a comprehensive WSI case bank capturing multi-scale morphological features with explicit reasoning pathways for experiential survival prediction.

\subsubsection{Gene‑Stratified CoT-Enhanced Case Bank}

For multimodal data integration, we introduce a Gene-Stratified CoT-Enhanced Case Bank to systematically analyze genomic data. Since raw genomic data is highly abstract with individual genes often lacking direct clinical value, we classify genes by functional roles into six prognostically important categories: Tumor Suppressor Genes, Oncogenes, Protein Kinases, Cell Differentiation Markers, Transcription Factors, and Cytokines and Growth Factors. Given genomic data $\mathcal{G}$, let $\mathcal{G}_l$ denote the gene subset of type $l$ where $l = \{1, \ldots, L\}, (L=6)$. For each type, we compute statistical features $\mathbf{s}_l = [\mu_l, m_l, r_{\text{mut}}^l]$, where $\mu_l$, $m_l$, and $r_{\text{mut}}^l$ represent mean expression, median, and mutation ratio, providing comprehensive quantitative characterization of expression and mutation patterns.

These statistics are submitted to GenAgent $\mathcal{A}_{\text{gen}}$ for preliminary analysis. GenAgent analyzes the statistical information with gene knowledge base $\mathcal{K}_{\text{gene}}$~\cite{xin2016high} to autonomously select genes with significant prognostic impact,
$\mathcal{G}^l = \mathcal{A}_{\text{gen}}^{\text{select}}(\mathbf{s}_l, \mathcal{G}_l; \mathcal{K}_{\text{gene}}),$
where $\mathcal{G}^*_l \subset \mathcal{G}_l$ contains selected important genes. GenAgent then retrieves raw expression data and consults $\mathcal{K}_{\text{gene}}$ to understand each gene's implications. Through coarse-to-fine analysis from statistics to gene details, GenAgent produces type-specific reports,
$\mathcal{R}_l^{\text{gene}} = \mathcal{A}_{\text{gen}}^{\text{report}}(\mathbf{s}_l, \mathcal{G}^*l; \mathcal{K}_{\text{gene}}).$

\begin{table*}[t]
\centering
\caption{Comparison of survival prediction performance (C-index) across different models and modalities on five TCGA cancer cohorts. G: Genomic modality, H: Histopathology modality. ``*" indicate best results from our reimplementation; Others are from original publications.}
\label{tab:survival_results}
\small
\scalebox{0.92}{
\begin{tabular}{lccccccc}
\toprule
\textbf{Model} & \textbf{Modality} & \textbf{BLCA} & \textbf{BRCA} & \textbf{GBMLGG} & \textbf{LUAD} & \textbf{UCEC} & \textbf{Overall} \\
 & & \textbf{(N = 373)} & \textbf{(N = 956)} & \textbf{(N = 480)} & \textbf{(N = 569)} & \textbf{(N = 453)} & \\
\midrule
\rowcolor{orange! 20}\multicolumn{8}{c}{Conventional Methods}\\
SNN*~\cite{klambauer2017self} & G & 0.541$\pm$0.016 & 0.466$\pm$0.058 & 0.598$\pm$0.054 & 0.539$\pm$0.069 & 0.493$\pm$0.096 & 0.527 \\
SNNTrans~\cite{zhou2024cohort} & G & 0.646$\pm$0.043 & 0.648$\pm$0.058 & 0.828$\pm$0.016 & 0.634$\pm$0.049 & 0.632$\pm$0.032 & 0.678 \\
AttnMIL~\cite{zhou2024cohort,ilse2018attentionbaseddeepmultipleinstance} & H & 0.605$\pm$0.045 & 0.551$\pm$0.077 & 0.816$\pm$0.011 & 0.563$\pm$0.050 & 0.614$\pm$0.052 & 0.630 \\
MaxMIL~\cite{zhou2024cohort} & H & 0.551$\pm$0.032 & 0.597$\pm$0.055 & 0.714$\pm$0.057 & 0.596$\pm$0.060 & 0.563$\pm$0.055 & 0.604 \\
DeepAttnMISL~\cite{Yao_2020,xu2023multimodal} & H & 0.504$\pm$0.042 & 0.524$\pm$0.043 & 0.734$\pm$0.029 & 0.548$\pm$0.050 & 0.597$\pm$0.059 & 0.581 \\
M3IF~\cite{zhou2024cohort,10.1007/978-3-030-87237-3_51} & G+H & 0.636$\pm$0.020 & 0.620$\pm$0.071 & 0.824$\pm$0.017 & 0.630$\pm$0.031 & 0.667$\pm$0.029 & 0.675 \\
HFBSurv~\cite{zhou2024cohort} & G+H & 0.640$\pm$0.028 & 0.647$\pm$0.035 & 0.838$\pm$0.013 & 0.650$\pm$0.050 & 0.642$\pm$0.045 & 0.683 \\
MOTCat*~\cite{xu2023multimodal} & G+H & \underline{0.674$\pm$0.024} & \underline{0.684$\pm$0.011} & 0.831$\pm$0.028 & \underline{0.674$\pm$0.036} & 0.667$\pm$0.051 & \underline{0.706} \\
MCAT*~\cite{Chen_2021_ICCV} & G+H & 0.645$\pm$0.053 & 0.601$\pm$0.069 & \textbf{0.852$\pm$0.028} & 0.636$\pm$0.043 & 0.634$\pm$0.018 & 0.674 \\
CCL*~\cite{zhou2024cohort} & G+H & 0.652$\pm$0.034 & 0.593$\pm$0.058 & \underline{0.845$\pm$0.012} & 0.640$\pm$0.059 & \underline{0.668$\pm$0.041} & 0.680 \\\hline
\rowcolor{green! 10}\multicolumn{8}{c}{Proprietary MLLMs} \\
Gemini-2.5-Pro*~\cite{google_gemini_2_5_pro_2025} & G+H & 0.572$\pm$0.031 & 0.555$\pm$0.055 & 0.551$\pm$0.026 & 0.531$\pm$0.062 & 0.498$\pm$0.067 & 0.541 \\
Claude-4.5*~\cite{anthropic_claude_4_5_2025} & G+H & 0.545$\pm$0.027 & 0.555$\pm$0.046 & 0.505$\pm$0.053 & 0.509$\pm$0.059 & 0.479$\pm$0.034 & 0.519 \\
GPT-5*~\cite{openai_gpt5_2025} & G+H & 0.576$\pm$0.038 & 0.434$\pm$0.057 & 0.493$\pm$0.053 & 0.510$\pm$0.087 & 0.495$\pm$0.083 & 0.502 \\\hline
\rowcolor{pink! 20}\multicolumn{8}{c}{Medical Agents} \\
MDAgent*~\cite{kim2024mdagents} & G+H & 0.558$\pm$0.040 & 0.482$\pm$0.064 & 0.495$\pm$0.040 & 0.524$\pm$0.064 & 0.509$\pm$0.049 & 0.514 \\
MedAgent*~\cite{tang2024medagentslargelanguagemodels} & G+H & 0.515$\pm$0.039 & 0.510$\pm$0.031 & 0.483$\pm$0.020 & 0.485$\pm$0.050 & 0.551$\pm$0.066 & 0.509 \\
\textbf{SurvAgent (Ours)} & G+H & \textbf{0.683$\pm$0.022} & \textbf{0.695$\pm$0.013} & 0.833$\pm$0.029 & \textbf{0.676 ± 0.036} & \textbf{0.676$\pm$0.052} & \textbf{0.713} \\
\bottomrule
\end{tabular}}
\end{table*}

After analyzing all six categories, GenAgent generates a comprehensive genomic report $\mathcal{R}_{\text{sum}}^{\text{gene}}$. Similar to WSI analysis, GenAgent generates CoT reasoning from the genomic report and GT patient survival class and time:
$\text{CoT}_{\text{gene}} = \mathcal{A}_{\text{gen}}(\mathcal{R}_{\text{sum}}^{\text{gene}}, t_{\text{GT}})$,
followed by self-critique and refinement:
\begin{equation}
\text{CoT}_{\text{gene}}^{\text{refined}} = \begin{cases}
\text{CoT}_{\text{gene}} & \text{if } \mathcal{V}(\text{CoT}_{\text{gene}}) = \text{high} \\
\mathcal{A}_{\text{gen}}(\text{CoT}_{\text{gene}}, \text{Critique}) & \text{Otherwise}
\end{cases}.
\end{equation}
The triplet $(\mathcal{R}_{\text{sum}}^{\text{gene}}, \text{CoT}_{\text{gene}}^{\text{refined}}, t_{\text{GT}})$ is stored in Gene CoT Case Bank $\mathcal{B}_{\text{gene}}$ for inference-time retrieval. Through this gene-stratified pipeline, from statistical characterization to targeted gene selection and CoT-enhanced analysis, we construct a comprehensive genomic case bank capturing functional genomic patterns with explicit reasoning pathways for knowledge-guided survival prediction

\subsection{Dichotomy-Based Multi-Expert Agent Inference Stage}

For a test case with WSI $\mathcal{W}_{\text{test}}$ and genomic data $\mathcal{G}_{\text{test}}$, we first generate multimodal reports using the same hierarchical pipeline. For WSI analysis, we apply the hierarchical analysis pipeline including LMScreen, CoSMining, and ConfMining to extract hierarchal WSI reports $\mathcal{R}_{\text{test}}^{\text{WSI}}$.
For genomic data, we perform gene-stratified analysis across six gene types to characterize genomic reports $\mathcal{R}_{\text{test}}^{\text{gene}}$ .

Using retrieval-augmented generation (RAG), we retrieve $K$ similar cases from both case banks based on multimodal report similarity:
\begin{equation}
\mathcal{B}_{\text{retrieved}} = \text{RAG}(\mathcal{R}_{\text{test}}^{\text{WSI}}, \mathcal{R}_{\text{test}}^{\text{gene}}; \mathcal{B}_{\text{WSI}}, \mathcal{B}_{\text{gene}}, K),
\end{equation}
where each retrieved case contains its summarized reports, CoT reasoning, and survival class and time.

Additionally, we collect predictions from $M$ expert survival models~\cite{xu2023multimodal,chen2021multimodal,zhou2024cohort}:
$\{\hat{t}_m\}_{m=1}^M = \{\mathcal{M}_m(\mathcal{W}_{\text{test}}, \mathcal{G}_{\text{test}})\}_{m=1}^M$,
where $\mathcal{M}_m$ represents the $m$-th expert model.
The inference agent $\mathcal{A}_{\text{infer}}$ performs dichotomy-based reasoning by progressively refining survival intervals. We define a hierarchical interval structure with $D$ dichotomy levels. At the first level, the agent performs coarse classification:
\begin{equation}y_1 = \mathcal{A}_{\text{infer}}(\mathcal{B}_{\text{retrieved}}, \mathcal{R}_{\text{test}}^{\text{WSI}}, \mathcal{R}_{\text{test}}^{\text{gene}}, \{\hat{t}_m\}; \text{level}=1),
\end{equation}
where $y_1 \in \{1, 2\}$ divides cases into two broad survival categories. At each subsequent level $d = \{ 2, \ldots, D \}$, the agent further refines the classification within the selected interval:
\begin{equation}
y_d = \mathcal{A}_{\text{infer}}(\mathcal{B}_{\text{retrieved}}, \mathcal{R}_{\text{test}}^{\text{WSI}}, \mathcal{R}_{\text{test}}^{\text{gene}}, \{\hat{t}_m\}, y_{d-1}; \text{level}=d),
\end{equation}
progressively narrowing the survival interval until reaching the finest granularity.
Finally, the exact survival time is predicted within the identified interval:
\begin{equation}
\hat{t}_{\text{final}} = \mathcal{A}_{\text{infer}}(\mathcal{B}_{\text{retrieved}}, \mathcal{R}_{\text{test}}^{\text{WSI}}, \mathcal{R}_{\text{test}}^{\text{gene}}, \{\hat{t}_m\}, y_D).
\end{equation}
The inference agent outputs comprehensive results including the predicted survival time $\hat{t}_{\text{final}}$, multimodal reports $(\mathcal{R}_{\text{test}}^{\text{WSI}}, \mathcal{R}_{\text{test}}^{\text{gene}})$, and an inference reasoning report $\mathcal{R}_{\text{reasoning}}$ that documents the complete decision-making process, providing transparency for clinical validation.

\section{Experiments}

\begin{table*}[t]
\centering
\caption{Ablation study of different components in \ours~framework.}
\label{tab:ablation}
\small
\scalebox{0.85}{
\begin{tabular}{ccc c c c c c c} 
\toprule
\textbf{WSI CoT bank} & \textbf{Gene CoT bank} & \textbf{Inference} & \textbf{BLCA} & \textbf{BRCA} & \textbf{GBMLGG} & \textbf{LUAD} & \textbf{UCEC} & \textbf{Overall} \\
\midrule
 &  &  & 0.452$\pm$0.030 & 0.421$\pm$0.033 & 0.463$\pm$0.024 & 0.498$\pm$0.086 & 0.470$\pm$0.045 & 0.461 \\
\checkmark &  &  & 0.612$\pm$0.053 & 0.542$\pm$0.095 & 0.791$\pm$0.024 & 0.559$\pm$0.039 & 0.585$\pm$0.050 & 0.618 \\
 & \checkmark &  & 0.539$\pm$0.016 & 0.455$\pm$0.056 & 0.591$\pm$0.050 & 0.545$\pm$0.073 & 0.481$\pm$0.082 & 0.522 \\
 &  & \checkmark & 0.664$\pm$0.041	& 0.665$\pm$0.012	& 0.813$\pm$0.024	& 0.650$\pm$0.039	& 0.652$\pm$0.052	& 0.689 \\
\checkmark & \checkmark & \checkmark & \textbf{0.683$\pm$0.022} & \textbf{0.695$\pm$0.013} & \textbf{0.833$\pm$0.029} & \textbf{0.676$\pm$0.033} & \textbf{0.676$\pm$0.052} & \textbf{0.713} \\
\bottomrule
\end{tabular}}
\end{table*}

\begin{figure*}[t]
    \centering
    \includegraphics[width=0.95\textwidth]{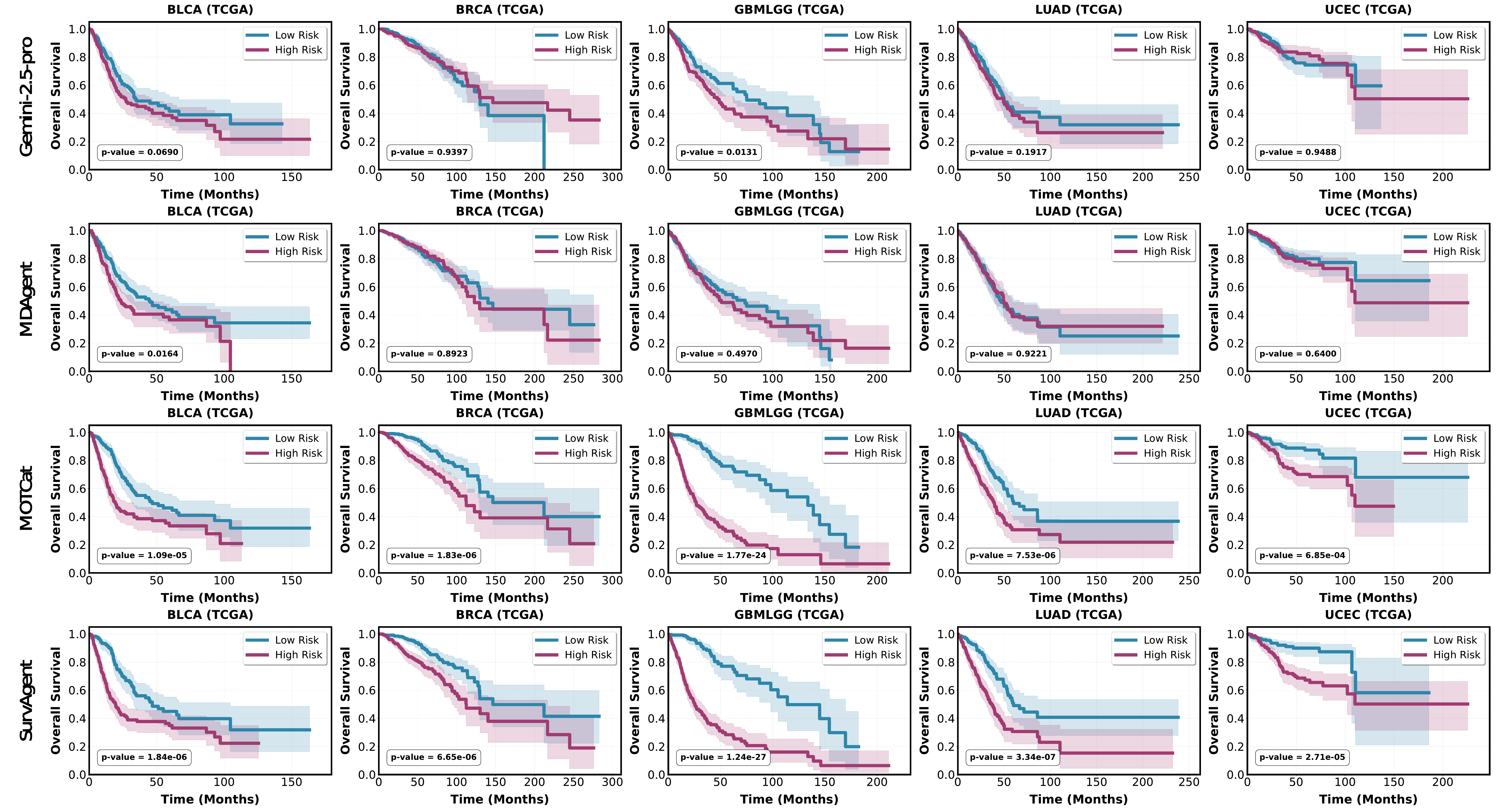}
    \caption{Kaplan-Meier Analysis of predicted high-risk (red) and low-risk (blue) groups on five cancer datasets and their p-values. Shaded areas refer to the confidence intervals.}
    \label{fig:km}
\end{figure*}

\subsection{Datasets and Settings}

\noindent\textbf{Datasets and Evaluation.} To ensure fair comparison, we follow prior protocols using five-fold cross-validation on five datasets: Bladder Urothelial Carcinoma (BLCA), Breast Invasive Carcinoma (BRCA), Glioblastoma and Lower Grade Glioma (GBMLGG), Lung Adenocarcinoma (LUAD), and Uterine Corpus Endometrial Carcinoma (UCEC). Data volumes remain consistent across datasets (details in Table~\ref{tab:survival_results}). We evaluate using the Concordance Index (C-index)\cite{harrell1996multivariable}, Kaplan-Meier survival curves\cite{kaplan1958nonparametric}, and Log-rank test~\cite{mantel1966evaluation} to assess survival differences between risk groups and validate prediction reliability.

\noindent\textbf{Implementation.} For each WSI, tissue regions are segmented via Otsu's thresholding. Non-overlapping $256\times256$ patches are extracted from tissue areas at $20\times$ magnification and processed with CLAM~\cite{lu2021data} to extract features for survival prediction models. Genomic data uses an SNN~\cite{klambauer2017self} encoder. CoSMining parameters: $\tau_v=\tau_t=0.93$; RAG: $K=3$; inference: $D=2$. Code will be publicly released.


\begin{figure*}[t]
    \centering
    \includegraphics[width=\textwidth]{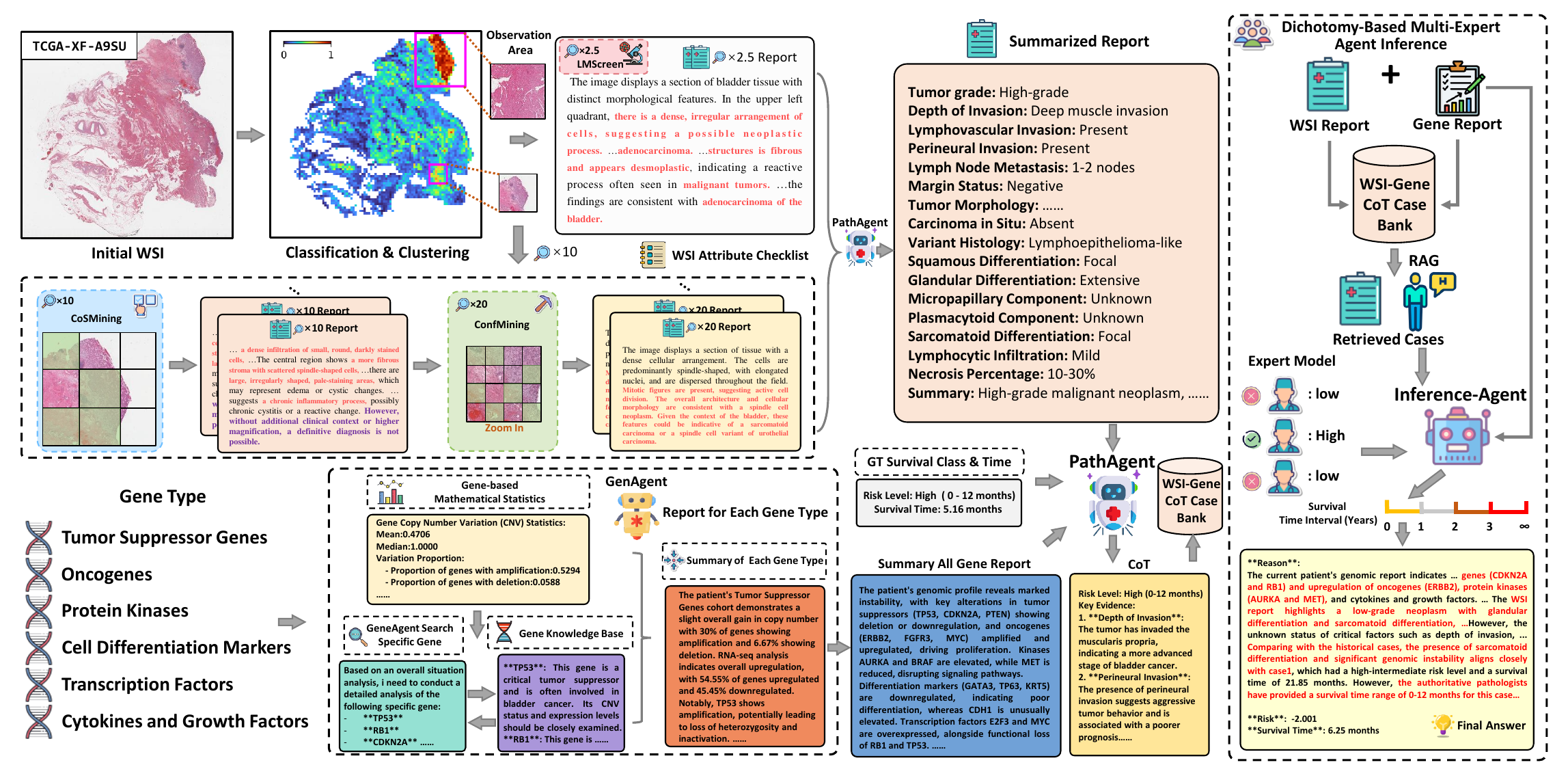}
    \caption{Explainability analysis through visualization on case TCGA-XF-A9SU}
    \label{fig:case_study_framework}
\end{figure*}

\subsection{Comparison with State-of-the-Arts}

To prove the superiority of our \ours, we compare it against SOTA approaches across five TCGA cancer cohorts in survival prediction, including conventional methods (unimodal and multimodal), leading proprietary MLLMs (Gemini 2.5 Pro~\cite{google_gemini_2_5_pro_2025}, Claude 4.5~\cite{anthropic_claude_4_5_2025}, GPT-5~\cite{openai_gpt5_2025}), and medical multi-agent systems (MedAgent~\cite{tang2024medagentslargelanguagemodels} and MDAgent~\cite{kim2024mdagents}), as shown in Table~\ref{tab:survival_results}. 

\noindent\textbf{Comparison with Conventional Methods.}
Our \ours~demonstrates superior performance over conventional survival prediction methods across most cohorts. Compared to the best conventional baseline MOTCat, \ours~achieves improvements of 0.9\%, 1.1\%, and 0.2\% in C-index on BLCA, BRCA, and LUAD, respectively, with 0.7\% overall gain. 
Compared to unimodal approaches, \ours~significantly surpasses genomic-based (SNNTrans: 0.678) and histopathology-based methods (AttnMIL: 0.630). 
The key advantage lies in our hierarchical WSI analysis and CoT-enhanced case banking enabling experiential learning, unlike conventional feature fusion methods, leading to more robust predictions.

\noindent\textbf{Comparison with Proprietary MLLMs.}
General-purpose proprietary MLLMs exhibit poor performance on survival prediction despite advanced capabilities elsewhere. Our \ours~substantially outperforms Gemini-2.5-Pro, Claude-4.5, and GPT-5 by 17.2\%, 19.4\%, and 21.1\% in overall C-index. On GBMLGG, the gap is most pronounced: \ours~achieves 0.833 versus Gemini-2.5-Pro (0.551), Claude-4.5 (0.505), and GPT-5 (0.493). Even where proprietary MLLMs perform relatively better, such as Gemini-2.5-Pro on BLCA (0.572), \ours~achieves 11.1\% gains. 
\xh{These results show that general MLLMs struggle with medical prediction tasks requiring domain knowledge and structured multimodal reasoning, highlighting the need for specialized models like \ours.
}

\noindent\textbf{Comparison with Medical Multi-Agent Systems.}
Our \ours~significantly outperforms existing medical multi-agent systems, achieving 19.9\% and 20.4\% overall C-index improvements over MDAgent and MedAgent. Across all cohorts, \ours~consistently demonstrates superior performance: improvements range from 12.5\% (BLCA) to 33.8\% (GBMLGG) over MDAgent, and 16.8\% (BLCA) to 35.0\% (GBMLGG) over MedAgent. These substantial gaps highlight advantages of our task-specific design: modality-specific agents, CoT-enhanced case banking for experiential learning, and dichotomy-based inference for transparent decisions, demonstrating that domain-specific multi-agent architectures significantly outperform general-purpose medical agents in complex clinical prediction.

\subsection{Ablation Study}

To evaluate the effectiveness of each proposed component, we conduct ablation studies by systematically removing individual modules. In Table~\ref{tab:ablation}, \ours~comprises three main components: the Hierarchical WSI CoT-Enhanced Case Bank, the Gene-Stratified CoT-Enhanced Case Bank, and the Dichotomy-Based Multi-Expert Agent Inference stage. When integrating only the WSI CoT bank with baseline survival models, the overall C-index improves from 0.461 to 0.618, showing the substantial value of hierarchical pathological case-based reasoning. Similarly, incorporating only the Gene CoT bank yields an overall C-index of 0.522, confirming that genomic experiential learning also contributes to performance improvement. When incorporating the proposed inference pipeline, the performance of baseline model boosts by 0.22 in C-index, suggesting its effectiveness. 
The complete \ours~framework, which integrates both case banks with dichotomy-based multi-agent inference, achieves the highest overall C-index of 0.713, demonstrating the synergistic benefits of multimodal case-based reasoning and progressive interval refinement.

\subsection{Patient Stratification}
Beyond C-index, patient stratification into distinct risk subgroups is critical for cancer prognosis. We compared SurvAgent with top models from each category: MOTCat (conventional), Gemini-2.5-pro (MLLM), and MDAgent (multi-agent) using Kaplan-Meier curves (Fig.~\ref{fig:km}).
MLLMs and multi-agent frameworks failed to discriminate high-risk from low-risk populations effectively. Gemini-2.5-pro and MDAgent showed non-significant results (p$>$0.05) on 80\% (4/5) of datasets, indicating unstable feature-outcome associations. SurvAgent achieved 100\% statistical significance (all p$<$0.05). On GBMLGG, SurvAgent demonstrated exceptional stratification (p=1.24e-27).
Compared to MOTCat, SurvAgent achieved comparable or better p-values, 
demonstrating its ability to leverage WSI and genomic features for robust survival prediction. 
This performance derives from our WSI-Gene CoT-Enhanced Case Bank and Dichotomy-Based Multi-Agent Inference, which together enable interpretable, accurate, and progressively refined risk stratification.
\subsection{Explainability Analysis}
To demonstrate the explainability and accuracy of \ours, we present a comprehensive case study in the Fig.1 for case TCGA-XF-A9SU. The hierarchical WSI analysis begins with LMScreen identifying adenocarcinoma with ``dense, irregular arrangement of cells" and desmoplastic structures. CoSMining's ×10 report reveals "dense infiltration of small, round, darkly stained cells" with fibrous stroma and spindle-shaped cells, noting ``without additional clinical context or higher magnification, a definitive diagnosis is not possible." ConfMining's ×20 report then identifies critical features: ``predominantly spindle-shaped cells with elongated nuclei" and ``mitotic figures present," with morphology ``consistent with spindle cell neoplasm indicative of sarcomatoid carcinoma``—prompting higher magnification analysis. The WSI Attribute Checklist extracts 15 features including deep muscle invasion, perineural invasion, and focal sarcomatoid differentiation, summarizing as ``high-grade malignant neoplasm with perineural invasion and necrosis." GenAgent analyzes genes (TP53, RB1, CDKN2A), revealing ``TP53 amplification potentially leading to loss of heterozygosity," ``marked instability with tumor suppressors (TP53, CDKN2A, PTEN) deletion/downregulation, oncogenes (ERBB2, FGFR3, MYC) amplified/upregulated," and ``overexpressed E2F3/MYC with functional loss of RB1/TP53." The CoT reasoning documents muscularis propria invasion and perineural invasion as aggressive indicators, classifying as high-risk (0-12 months) the same as the GT low-risk labeling (5.16 months). 
The final inference agent reconciles conflicting signals—e.g., ``sarcomatoid differentiation and genomic instability align with case1 (21.85 months), but experts estimate 0–12 months''—and through dichotomy-based reasoning predicts 6.25 months (risk = –2.001) versus the ground truth 5.16 months, transparently balancing contradictory evidence for interpretable clinical reasoning.


\section{Conclusion}
We present \ours, the first multi-agent system for multimodal survival prediction. Through WSI-Gene CoT-Enhanced Case Banking, we enable experiential learning by storing analytical reasoning processes from historical cases. Our hierarchical pipeline balances accuracy, efficiency, and coverage via LMScreen, CoSMining, and ConfMining, while dichotomy-based inference ensures transparent predictions. Extensive experiments show our superiority to conventional methods, MLLMs, and medical agents with consistent statistical significance in patient stratification. 


{
    \small
    \bibliographystyle{ieeenat_fullname}
    \bibliography{main}

@String(ICCV= {Int. Conf. Comput. Vis.})

@String(ICCV  = {ICCV})

@article{lou2025cxragent,
  title={CXRAgent: Director-Orchestrated Multi-Stage Reasoning for Chest X-Ray Interpretation},
  author={Lou, Jinhui and Yang, Yan and Yu, Zhou and Fu, Zhenqi and Han, Weidong and Huang, Qingming and Yu, Jun},
  journal={arXiv preprint arXiv:2510.21324},
  year={2025}
}

@article{tzanis2025agentic,
  title={Agentic systems in radiology: Principles, opportunities, privacy risks, regulation, and sustainability concerns},
  author={Tzanis, Eleftherios and Adams, Lisa C and D’Antonoli, Tugba Akinci and Bressem, Keno K and Cuocolo, Renato and Kocak, Burak and Malamateniou, Christina and Klontzas, Michail E},
  journal={Diagnostic and Interventional Imaging},
  year={2025},
  publisher={Elsevier}
}

@article{tang2025endoagent,
  title={EndoAgent: A Memory-Guided Reflective Agent for Intelligent Endoscopic Vision-to-Decision Reasoning},
  author={Tang, Yi and Wang, Kaini and Chen, Yang and Zhou, Guangquan},
  journal={arXiv preprint arXiv:2508.07292},
  year={2025}
}

@article{ferber2025development,
  title={Development and validation of an autonomous artificial intelligence agent for clinical decision-making in oncology},
  author={Ferber, Dyke and El Nahhas, Omar SM and W{\"o}lflein, Georg and Wiest, Isabella C and others},
  journal={Nature cancer},
  pages={1--13},
  year={2025},
  publisher={Nature Publishing Group US New York}
}

@inproceedings{quang2025gmat,
  title={GMAT: Grounded Multi-agent Clinical Description Generation for Text Encoder in Vision-Language MIL for Whole Slide Image Classification},
  author={Quang, Ngoc Bui Lam and Binh, Nam Le Nguyen and Nguyen, Thanh-Huy and Nguyen, Le Thien Phuc and Nguyen, Quan and Bagci, Ulas},
  booktitle={International Workshop on Emerging LLM/LMM Applications in Medical Imaging},
  pages={1--9},
  year={2025},
  organization={Springer}
}

@article{wang2025pathology,
  title={Pathology-CoT: Learning Visual Chain-of-Thought Agent from Expert Whole Slide Image Diagnosis Behavior},
  author={Wang, Sheng and Wu, Ruiming and Herndon, Charles and Liu, Yihang and Koga, Shunsuke and Shen, Jeanne and Huang, Zhi},
  journal={arXiv preprint arXiv:2510.04587},
  year={2025}
}

@article{sun2025cpathagent,
  title={CPathAgent: An Agent-based Foundation Model for Interpretable High-Resolution Pathology Image Analysis Mimicking Pathologists' Diagnostic Logic},
  author={Sun, Yuxuan and Si, Yixuan and Zhu, Chenglu and Zhang, Kai and Shui, Zhongyi and Ding, Bowen and Lin, Tao and Yang, Lin},
  journal={arXiv preprint arXiv:2505.20510},
  year={2025}
}

@article{ghezloo2025pathfinder,
  title={Pathfinder: A multi-modal multi-agent system for medical diagnostic decision-making applied to histopathology},
  author={Ghezloo, Fatemeh and Seyfioglu, Mehmet Saygin and Soraki, Rustin and Ikezogwo, Wisdom O and Li, Beibin and Vivekanandan, Tejoram and Elmore, Joann G and Krishna, Ranjay and Shapiro, Linda},
  journal={arXiv preprint arXiv:2502.08916},
  year={2025}
}

@article{chen2025evidence,
  title={Evidence-based diagnostic reasoning with multi-agent copilot for human pathology},
  author={Chen, Chengkuan and Weishaupt, Luca L and Williamson, Drew FK and Chen, Richard J and others},
  journal={arXiv preprint arXiv:2506.20964},
  year={2025}
}

@inproceedings{lyu2025inline,
  title={WSI-Agents: A Collaborative Multi-agent System for Multi-modal Whole Slide Image Analysis},
  author={Lyu, Xinheng and Liang, Yuci and Chen, Wenting and Ding, Meidan and Yang, Jiaqi and Huang, Guolin and Zhang, Daokun and He, Xiangjian and Shen, Linlin},
  booktitle={International Conference on Medical Image Computing and Computer-Assisted Intervention},
  pages={680--690},
  year={2025}
}

@article{li2025care,
  title={CARE-AD: a multi-agent large language model framework for Alzheimer’s disease prediction using longitudinal clinical notes},
  author={Li, Rumeng and Wang, Xun and Berlowitz, Dan and Mez, Jesse and Lin, Honghuang and Yu, Hong},
  journal={npj Digital Medicine},
  volume={8},
  number={1},
  pages={541},
  year={2025},
  publisher={Nature Publishing Group UK London}
}

@article{li2024agent,
  title={Agent hospital: A simulacrum of hospital with evolvable medical agents},
  author={Li, Junkai and Lai, Yunghwei and Li, Weitao and Ren, Jingyi and others},
  journal={arXiv preprint arXiv:2405.02957},
  year={2024}
}

@article{jiang2024long,
  title={Long term memory: The foundation of ai self-evolution},
  author={Jiang, Xun and Li, Feng and Zhao, Han and Qiu, Jiahao and Wang, Jiaying and others},
  journal={arXiv preprint arXiv:2410.15665},
  year={2024}
}

@article{lu2021data,
  title={Data-efficient and weakly supervised computational pathology on whole-slide images},
  author={Lu, Ming Y and Williamson, Drew FK and Chen, Tiffany Y and Chen, Richard J and Barbieri, Matteo and Mahmood, Faisal},
  journal={Nature biomedical engineering},
  volume={5},
  number={6},
  pages={555--570},
  year={2021},
  publisher={Nature Publishing Group UK London}
}

@article{gyHorffy2021survival,
  title={Survival analysis across the entire transcriptome identifies biomarkers with the highest prognostic power in breast cancer},
  author={Gy{\H{o}}rffy, Bal{\'a}zs},
  journal={Computational and structural biotechnology journal},
  volume={19},
  pages={4101--4109},
  year={2021},
  publisher={Elsevier}
}

@inproceedings{jaume2024modeling,
  title={Modeling dense multimodal interactions between biological pathways and histology for survival prediction},
  author={Jaume, Guillaume and Vaidya, Anurag and Chen, Richard J and Williamson, Drew FK and Liang, Paul Pu and Mahmood, Faisal},
  booktitle={Proceedings of the IEEE Conf. Comput. Vis. Pattern Recognit.},
  pages={11579--11590},
  year={2024}
}

@inproceedings{zhou2023cross,
  title={Cross-modal translation and alignment for survival analysis},
  author={Zhou, Fengtao and Chen, Hao},
  booktitle={Proceedings of the IEEE Int. Conf. Comput. Vis.},
  pages={21485--21494},
  year={2023}
}

@article{zhou2024cohort,
  title={Cohort-Individual Cooperative Learning for Multimodal Cancer Survival Analysis},
  author={Zhou, Huajun and Zhou, Fengtao and Chen, Hao},
  journal={arXiv preprint arXiv:2404.02394},
  year={2024}
}

@article{zhang2024prototypical,
  title={Prototypical information bottlenecking and disentangling for multimodal cancer survival prediction},
  author={Zhang, Yilan and Xu, Yingxue and Chen, Jianqi and Xie, Fengying and Chen, Hao},
  journal={arXiv preprint arXiv:2401.01646},
  year={2024}
}

@inproceedings{song2024morphological,
  title={Morphological prototyping for unsupervised slide representation learning in computational pathology},
  author={Song, Andrew H and Chen, Richard J and Ding, Tong and Williamson, Drew FK and Jaume, Guillaume and Mahmood, Faisal},
  booktitle={Proceedings of the IEEE/CVF Conference on Computer Vision and Pattern Recognition},
  pages={11566--11578},
  year={2024}
}

@article{zhou2025robust,
  title={Robust multimodal survival prediction with the latent differentiation conditional variational autoencoder},
  author={Zhou, Junjie and Tang, Jiao and Zuo, Yingli and Wan, Peng and Zhang, Daoqiang and Shao, Wei},
  journal={arXiv preprint arXiv:2503.09496},
  year={2025}
}

@article{kim2024mdagents,
  title={Mdagents: An adaptive collaboration of llms for medical decision-making},
  author={Kim, Yubin and Park, Chanwoo and Jeong, Hyewon and Chan, Yik S and Xu, Xuhai and McDuff, Daniel and Lee, Hyeonhoon and Ghassemi, Marzyeh and Breazeal, Cynthia and Park, Hae W},
  journal={Advances in Neural Information Processing Systems},
  volume={37},
  pages={79410--79452},
  year={2024}
}

@article{liu2024medchain,
  title={Medchain: Bridging the gap between llm agents and clinical practice through interactive sequential benchmarking},
  author={Liu, Jie and Wang, Wenxuan and Ma, Zizhan and Huang, Guolin and SU, Yihang and Chang, Kao-Jung and Chen, Wenting and Li, Haoliang and Shen, Linlin and Lyu, Michael},
  journal={arXiv preprint arXiv:2412.01605},
  year={2024}
}

@article{donroe2024clinical,
  title={Clinical Reasoning: Perspectives of Expert Clinicians on Reasoning Through Complex Clinical Cases},
  author={Donroe, Joseph H and Egger, Emilie and Soares, Sarita and Sofair, Andre N and Moriarty, John},
  journal={Cureus},
  volume={16},
  number={1},
  year={2024},
  publisher={Cureus}
}

@article{rutter2025importance,
  title={The importance of clinical reasoning in differential diagnosis for non-medical prescribers, nurses and pharmacists},
  author={Rutter, Paul},
  journal={Clinics in Integrated Care},
  volume={31},
  pages={100271},
  year={2025},
  publisher={Elsevier}
}

@article{choi2023using,
  title={Using an experiential learning model to teach clinical reasoning theory and cognitive bias: an evaluation of a first-year medical student curriculum},
  author={Choi, Justin J and Gribben, Jeanie and Lin, Myriam and Abramson, Erika L and Aizer, Juliet},
  journal={Medical Education Online},
  volume={28},
  number={1},
  pages={2153782},
  year={2023},
  publisher={Taylor \& Francis}
}

@inproceedings{abbas2025explainable,
  title={Explainable AI in Clinical Decision Support Systems: A Meta-Analysis of Methods, Applications, and Usability Challenges},
  author={Abbas, Qaiser and Jeong, Woonyoung and Lee, Seung Won},
  booktitle={Healthcare},
  volume={13},
  number={17},
  pages={2154},
  year={2025},
  organization={MDPI}
}

@article{raz2025explainable,
  title={Explainable AI in medicine: challenges of integrating XAI into the future clinical routine},
  author={R{\"a}z, Tim and Pahud De Mortanges, Aur{\'e}lie and Reyes, Mauricio},
  journal={Frontiers in Radiology},
  volume={5},
  pages={1627169},
  year={2025},
  publisher={Frontiers}
}

@article{chen2020pathomic,
  title={Pathomic fusion: an integrated framework for fusing histopathology and genomic features for cancer diagnosis and prognosis},
  author={Chen, Richard J and Lu, Ming Y and Wang, Jingwen and Williamson, Drew FK and Rodig, Scott J and Lindeman, Neal I and Mahmood, Faisal},
  journal={IEEE Transactions on Medical Imaging},
  volume={41},
  number={4},
  pages={757--770},
  year={2020},
  publisher={IEEE}
}

@inproceedings{chen2021multimodal,
  title={Multimodal co-attention transformer for survival prediction in gigapixel whole slide images},
  author={Chen, Richard J and Lu, Ming Y and Weng, Wei-Hung and Chen, Tiffany Y and Williamson, Drew FK and Manz, Trevor and Shady, Maha and Mahmood, Faisal},
  booktitle={Proceedings of the IEEE/CVF international conference on computer vision},
  pages={4015--4025},
  year={2021}
}

@inproceedings{xu2023multimodal,
  title={Multimodal optimal transport-based co-attention transformer with global structure consistency for survival prediction},
  author={Xu, Yingxue and Chen, Hao},
  booktitle={Proceedings of the IEEE/CVF international conference on computer vision},
  pages={21241--21251},
  year={2023}
}

@article{wang2024pathology,
  title={A pathology foundation model for cancer diagnosis and prognosis prediction},
  author={Wang, Xiyue and Zhao, Junhan and Marostica, Eliana and Yuan, Wei and Jin, Jietian and others},
  journal={Nature},
  volume={634},
  number={8035},
  pages={970--978},
  year={2024},
  publisher={Nature Publishing Group UK London}
}

@inproceedings{devlin2019bert,
  title={Bert: Pre-training of deep bidirectional transformers for language understanding},
  author={Devlin, Jacob and Chang, Ming-Wei and Lee, Kenton and Toutanova, Kristina},
  booktitle={Proceedings of the 2019 conference of the North American chapter of the association for computational linguistics: human language technologies, volume 1 (long and short papers)},
  pages={4171--4186},
  year={2019}
}

@article{xin2016high,
  title={High-performance web services for querying gene and variant annotation},
  author={Xin, Jiwen and Mark, Adam and Afrasiabi, Cyrus and Tsueng, Ginger and Juchler, Moritz and others},
  journal={Genome biology},
  volume={17},
  number={1},
  pages={91},
  year={2016},
  publisher={Springer}
}

@misc{MyPathologyReport_zhTW,
title = {My Pathology Report | Patient Pathology (Traditional Chinese)},
author = {{MyPathologyReport Team}},
howpublished = {\url{https://www.mypathologyreport.ca/zh-TW/}},
note = {Internationally recognized, pathologist-written and peer-reviewed patient education resource; disclaimer: content is for general information, not individualized medical advice},
institution = {MyPathologyReport.ca},
year = {2025},
month = nov,
urldate = {2025-11-14}
}

@article{harrell1996multivariable,
  title={Multivariable prognostic models: issues in developing models, evaluating assumptions and adequacy, and measuring and reducing errors},
  author={Harrell Jr, Frank E and Lee, Kerry L and Mark, Daniel B},
  journal={Statistics in medicine},
  volume={15},
  number={4},
  pages={361--387},
  year={1996},
  publisher={Wiley Online Library}
}

@article{mantel1966evaluation,
  title={Evaluation of survival data and two new rank order statistics arising in its consideration},
  author={Mantel, Nathan and others},
  journal={Cancer Chemother Rep},
  volume={50},
  number={3},
  pages={163--170},
  year={1966}
}

@article{klambauer2017self,
  title={Self-normalizing neural networks},
  author={Klambauer, G{\"u}nter and Unterthiner, Thomas and Mayr, Andreas and Hochreiter, Sepp},
  journal={Advances in neural information processing systems},
  volume={30},
  year={2017}
}

@article{kaplan1958nonparametric,
  title={Nonparametric estimation from incomplete observations},
  author={Kaplan, Edward L and Meier, Paul},
  journal={Journal of the American statistical association},
  volume={53},
  number={282},
  pages={457--481},
  year={1958},
  publisher={Taylor \& Francis}
}

@misc{ilse2018attentionbaseddeepmultipleinstance,
      title={Attention-based Deep Multiple Instance Learning}, 
      author={Maximilian Ilse and Jakub M. Tomczak and Max Welling},
      year={2018},
      eprint={1802.04712},
      archivePrefix={arXiv},
      primaryClass={cs.LG},
      url={https://arxiv.org/abs/1802.04712}, 
}

@article{Yao_2020,
   title={Whole slide images based cancer survival prediction using attention guided deep multiple instance learning networks},
   volume={65},
   ISSN={1361-8415},
   url={http://dx.doi.org/10.1016/j.media.2020.101789},
   DOI={10.1016/j.media.2020.101789},
   journal={Medical Image Analysis},
   publisher={Elsevier BV},
   author={Yao, Jiawen and Zhu, Xinliang and Jonnagaddala, Jitendra and Hawkins, Nicholas and Huang, Junzhou},
   year={2020},
   month=oct, pages={101789} }

@InProceedings{10.1007/978-3-030-87237-3_51,
    author="Li, Hang
    and Yang, Fan
    and Xing, Xiaohan
    and Zhao, Yu
    and Zhang, Jun
    and Liu, Yueping
    and Han, Mengxue
    and Huang, Junzhou
    and Wang, Liansheng
    and Yao, Jianhua",
    editor="de Bruijne, Marleen
    and Cattin, Philippe C.
    and Cotin, St{\'e}phane
    and Padoy, Nicolas
    and Speidel, Stefanie
    and Zheng, Yefeng
    and Essert, Caroline",
    title="Multi-modal Multi-instance Learning Using Weakly Correlated Histopathological Images and Tabular Clinical Information",
    booktitle="Medical Image Computing and Computer Assisted Intervention -- MICCAI 2021",
    year="2021",
    publisher="Springer International Publishing",
    address="Cham",
    pages="529--539",
    isbn="978-3-030-87237-3"
}

@misc{tang2024medagentslargelanguagemodels,
      title={MedAgents: Large Language Models as Collaborators for Zero-shot Medical Reasoning}, 
      author={Xiangru Tang and Anni Zou and Zhuosheng Zhang and Ziming Li and Yilun Zhao and Xingyao Zhang and Arman Cohan and Mark Gerstein},
      year={2024},
      eprint={2311.10537},
      archivePrefix={arXiv},
      primaryClass={cs.CL},
      url={https://arxiv.org/abs/2311.10537}, 
}

@InProceedings{Chen_2021_ICCV,
    author    = {Chen, Richard J. and Lu, Ming Y. and Weng, Wei-Hung and Chen, Tiffany Y. and Williamson, Drew F.K. and Manz, Trevor and Shady, Maha and Mahmood, Faisal},
    title     = {Multimodal Co-Attention Transformer for Survival Prediction in Gigapixel Whole Slide Images},
    booktitle = {Proceedings of the IEEE/CVF International Conference on Computer Vision (ICCV)},
    month     = {October},
    year      = {2021},
    pages     = {4015-4025}
}

@software{openai_gpt5_2025,
  author       = {OpenAI},
  title        = {GPT-5 (Version 2025-08-07) [Large language model]},
  year         = {2025},
  url          = {https://openai.com/gpt-5},
  note         = {Accessed 2025-11-14}
}

@software{google_gemini_2_5_pro_2025,
  author= {Google DeepMind},
  title= {Gemini 2.5 Pro [Large language model]},
  year= {2025},
  url= {https://deepmind.google/technologies/gemini},
  note= {Accessed 2025-11-14}
}

@software{anthropic_claude_4_5_2025,
  author       = {Anthropic},
  title        = {Claude 4.5 [Large language model]},
  year         = {2025},
  url          = {https://www.anthropic.com/claude},
  note         = {Accessed 2025-11-14}
}

@misc{sun2024pathgen16m16millionpathology,
      title={PathGen-1.6M: 1.6 Million Pathology Image-text Pairs Generation through Multi-agent Collaboration}, 
      author={Yuxuan Sun and Yunlong Zhang and Yixuan Si and Chenglu Zhu and Zhongyi Shui and Kai Zhang and Jingxiong Li and Xingheng Lyu and Tao Lin and Lin Yang},
      year={2024},
      eprint={2407.00203},
      archivePrefix={arXiv},
      primaryClass={cs.CV},
      url={https://arxiv.org/abs/2407.00203}, 
}

@misc{qwen2025qwen25,
  title={Qwen2.5 Technical Report}, 
  author={Qwen Team and others},
  year={2025},
  eprint={2412.15115},
  archivePrefix={arXiv},
  primaryClass={cs.CL},
  url={https://arxiv.org/abs/2412.15115}, 
}

@misc{zhang2025pathoagenticragmultimodalagenticretrievalaugmented,
      title={Patho-AgenticRAG: Towards Multimodal Agentic Retrieval-Augmented Generation for Pathology VLMs via Reinforcement Learning}, 
      author={Wenchuan Zhang and Jingru Guo and Hengzhe Zhang and Penghao Zhang and Jie Chen and Shuwan Zhang and Zhang Zhang and Yuhao Yi and Hong Bu},
      year={2025},
      eprint={2508.02258},
      archivePrefix={arXiv},
      primaryClass={cs.CV},
      url={https://arxiv.org/abs/2508.02258}, 
}

@misc{deepseekv32025,
  title={DeepSeek-V3 Technical Report}, 
  author={DeepSeek-AI and others},
  year={2025},
  eprint={2412.19437},
  archivePrefix={arXiv},
  primaryClass={cs.CL},
  url={https://arxiv.org/abs/2412.19437}, 
}
}

\clearpage
\setcounter{page}{1}
\maketitlesupplementary
\appendix
\setcounter{table}{0}
\setcounter{figure}{0}


\noindent\textbf{Abstract.} 
Appendix A presents the performance comparison between \ours~and existing open-source pathology-specific multi-agent frameworks on the survival prediction task.
Appendix B provides the full WSI Attribute Checklist, including detailed medical definitions for each attribute and their relevance to patient prognosis.
Appendix C showcases two complete inference case studies, visualizing \ours’s step-by-step analysis over WSI and genomic data and its final reasoning process.
Appendix D details the structure and contents of the CoT Case Bank.
Appendix E includes all core prompt designs used in \ours.
Appendix F describes the implementation details and experimental settings used to construct and evaluate \ours. The source code will be released for public access.

\section{Comparison with Pathology Multi-Agent Framework}

\begin{table*}[t]
\centering
\caption{The survival prediction performance (C-index) of the specialized pathological multi-agent framework in five TCGA cancer research groups was compared. ``*" indicates the best results from our reimplementation.}
\label{tab:Pathology_Multi-Agent_Framework_performance}
\begin{tabular}{lcccccc}
\toprule
Model & BLCA & BRCA & GBMLGG & LUAD & UCEC & Overall \\
\midrule
WSI-Agent*~\cite{lyu2025inline} & \underline{0.566 $\pm$ 0.062} & \underline{0.568 $\pm$ 0.068} & 0.390 $\pm$ 0.030 & 0.518 $\pm$ 0.083 & \underline{0.577 $\pm$ 0.055} & \underline{0.524} \\
Patho-AgenticRAG*~\cite{zhang2025pathoagenticragmultimodalagenticretrievalaugmented}  & 0.478 $\pm$ 0.040 & 0.511 $\pm$ 0.083 & \underline{0.527 $\pm$ 0.063} & \underline{0.523 $\pm$ 0.059} & 0.507 $\pm$ 0.085 & 0.509 \\
SurvAgent         & \textbf{0.683 $\pm$ 0.022} & \textbf{0.695 $\pm$ 0.013} & \textbf{0.833 $\pm$ 0.029} & \textbf{0.676 $\pm$ 0.036} & \textbf{0.676 $\pm$ 0.052} & \textbf{0.713} \\
\bottomrule
\end{tabular}
\end{table*}

We evaluated \ours~against pathology-specific multi-agent frameworks, WSI-Agent~\cite{lyu2025inline} and Patho-AgenticRAG~\cite{zhang2025pathoagenticragmultimodalagenticretrievalaugmented}, across five TCGA cancer cohorts (Table~\ref{tab:Pathology_Multi-Agent_Framework_performance}). Existing frameworks demonstrate limited performance in survival prediction, with overall C-indexes of 0.524 and 0.509, and particularly low values on challenging cohorts such as GBMLGG (0.39 and 0.527). In contrast, \ours~consistently achieves superior prognostic accuracy across all datasets, with an overall C-index of 0.713, improving 19.65\% over existing methods and up to 0.833 on GBMLGG, representing a gain of 30.6\% over Patho-AgenticRAG (0.527). These results highlight that our multi-agent framework effectively integrates multi-modal pathological information and substantially outperforms specialized pathology agents in survival prediction tasks.

\section{WSI Attribute Checklist Overview}

\begin{figure}[t]
    \centering
    \includegraphics[width=\linewidth]{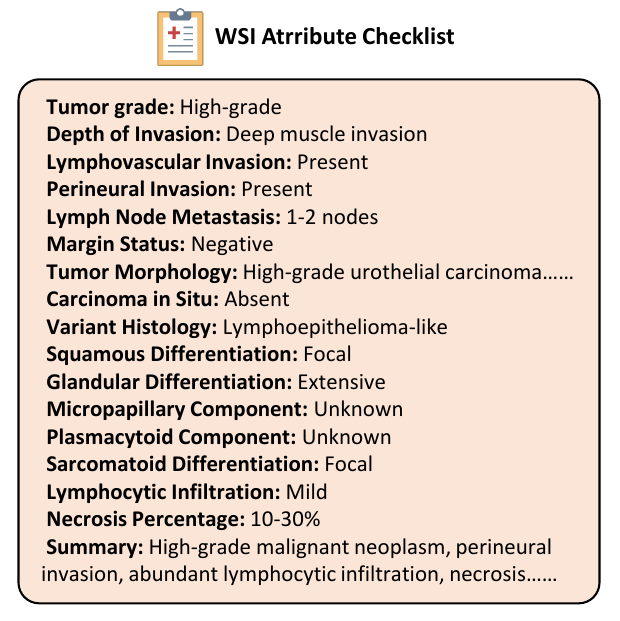}
    \caption{Overview of the WSI Attribute Checklist}
    \label{fig:wsi_atrribute_checklist}
\end{figure}

In this section, we provide a detailed illustration of the WSI Attribute Checklist used in \ours. Within our framework, whole slide images (WSIs) serve as one of the primary modalities for downstream survival prediction, and a central challenge lies in identifying which microscopic pathological characteristics most strongly influence patient prognosis. To address this, we construct a curated checklist consisting of 16 prognostic histopathological attributes, complemented by an additional global descriptor summarizing the overall characteristics of the current slide. All attributes are automatically extracted from WSIs by PathAgent and compiled by the Search Agent, which aggregates their definitions and clinical relevance based on pathology guidelines and established medical literature. An example of the full checklist is shown in Fig.~\ref{fig:wsi_atrribute_checklist}, and the meaning and definition of each attribute are described below.


\noindent\textbf{Tumor Grade.} A histological assessment of the tumor's cellular abnormality and proliferation rate. Reflects the intrinsic biological aggressiveness and growth potential of the tumor. A powerful prognostic indicator; high-grade tumors are consistently associated with significantly higher risks of recurrence and progression to invasive disease, whereas low-grade tumors typically follow a more indolent clinical course.

\noindent\textbf{Depth of Invasion.} The extent to which the tumor infiltrates the bladder wall, ranging from non-invasive mucosal involvement to deep muscular penetration.Reflects the intrinsic aggressiveness and progression stage of the tumor. One of the strongest predictors of survival; deeper invasion is consistently associated with markedly worse overall survival.

\noindent\textbf{Lymphovascular Invasion (LVI).} Presence of tumor cells within lymphatic or vascular channels. Indicates early metastatic potential. LVI is a well-established independent risk factor for distant metastasis and shortened survival.

\noindent\textbf{Perineural Invasion (PNI).} Tumor infiltration along or around nerve fibers. Suggests highly invasive tumor behavior. Strongly associated with local recurrence and poor long-term outcomes.

\noindent\textbf{Lymph Node Metastasis.} Involvement and number of metastatic regional lymph nodes. Forms a crucial component of TNM staging. Increasing nodal burden correlates with stepwise reduction in survival probability.

\noindent\textbf{Margin Status.} Presence (positive) or absence (negative) of residual tumor at the surgical resection margin. Positive margins imply incomplete tumor removal. Strong predictor of local relapse and reduced survival following surgery.

\noindent\textbf{Tumor Morphology.} Serves as the foundational visual evidence for tumor grading and subtyping. The morphological assessment is critical; high-grade features, including nuclear pleomorphism and frequent mitotic figures, are directly correlated with aggressive clinical behavior, and the identification of specific variant morphologies (e.g., micropapillary, sarcomatoid) carries significant prognostic and therapeutic implications.

\noindent\textbf{Carcinoma in Situ (CIS).} A high-grade non-invasive flat lesion with strong malignant potential. Often coexists with aggressive invasive disease. The presence (especially extensive) of CIS predicts increased progression risk.

\noindent\textbf{Variant Histology.} Non-conventional morphological subtypes such as squamous, glandular, micropapillary, sarcomatoid, plasmacytoid, nested, or lymphoepithelioma-like patterns. These variants frequently exhibit distinct biological behaviors. Many variants, particularly micropapillary, plasmacytoid, and sarcomatoid types, are associated with highly aggressive disease and poor survival.

\noindent\textbf{Squamous/Glandular Differentiation.} Partial or extensive differentiation toward squamous or glandular phenotypes. Represents divergent tumor evolution. Extensive differentiation is associated with advanced disease and inferior clinical outcomes.

\noindent\textbf{Micropapillary, Plasmacytoid, and Sarcomatoid Components.} Distinct morphologic components reflecting specific aggressive histologic variants. Indicate profound alterations in tumor microarchitecture. These components are widely recognized as markers of extremely poor prognosis.

\noindent\textbf{Lymphocytic Infiltration.} Degree of immune cell infiltration within the tumor microenvironment (TME). Reflects host immune response. Higher infiltration levels often correlate with more favorable outcomes, whereas minimal infiltration suggests an “immune-cold” phenotype.

\noindent\textbf{Necrosis Percentage.} Proportion of necrotic tumor areas. Indicates rapid tumor cell turnover and insufficient vascular supply. Extensive necrosis is a known indicator of aggressive tumor biology and poor survival.

\noindent\textbf{Summary.} A concise, free-text summarization that provides a comprehensive characterization of the current WSI intended to mitigate limitations of fixed-value attribute sets. Because the checklist attributes are discretized and selected by prior feature-filtering steps, they may omit subtle, rare, or composite histopathological cues and are susceptible to selection or annotation bias. The Summary Attribute is therefore designed to (1) capture additional prognostic signals not well represented by the predefined categorical fields, (2) record observations where multiple features interact or where uncertainty exists, and (3) serve as a corrective, interpretability-focused descriptor that complements the structured attributes for downstream risk stratification.

\section{Case Study Examples of \ours’s Reasoning Process}

\begin{figure*}[!t]
    \centering
    \includegraphics[width=\textwidth]{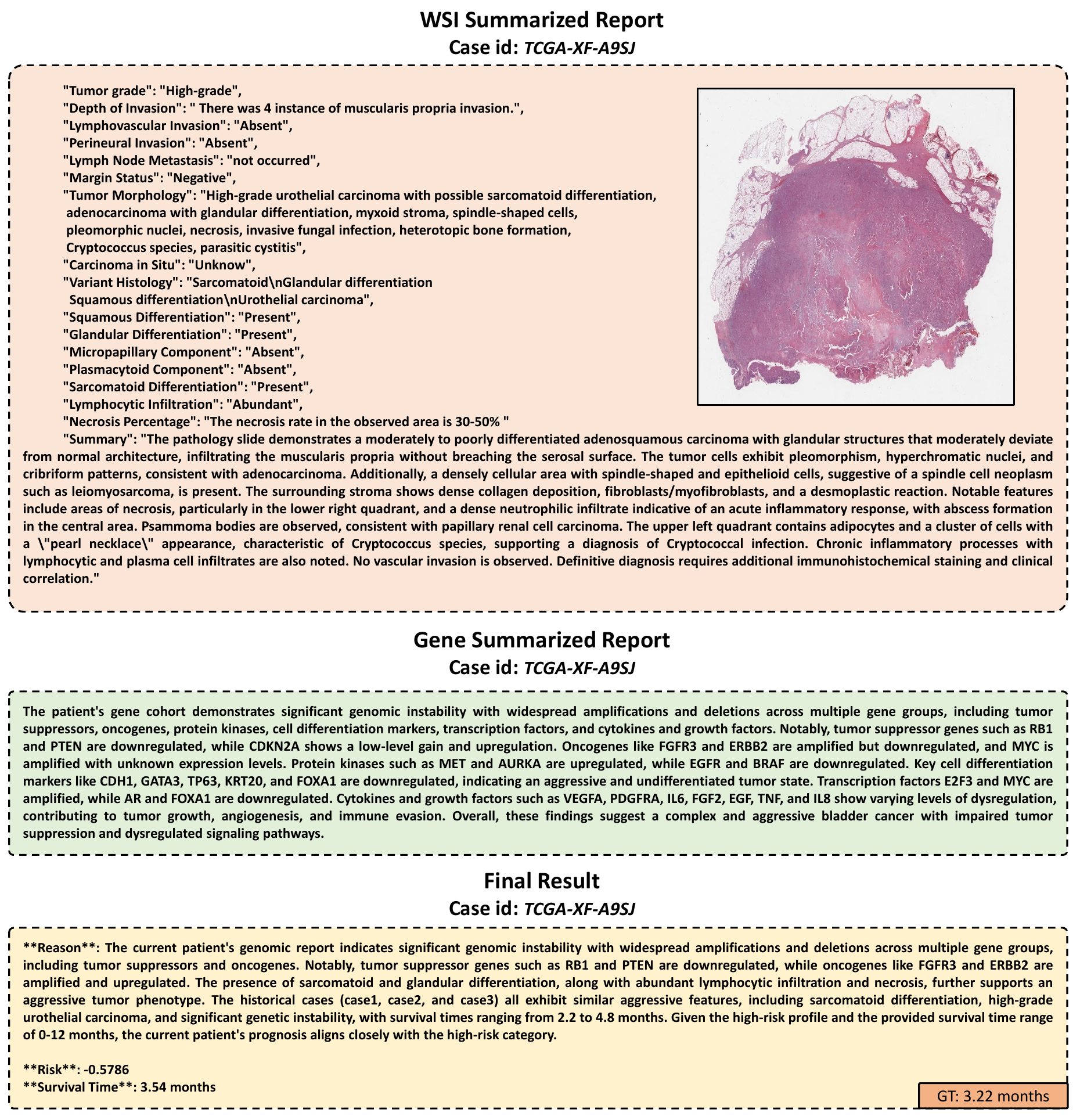}
    \caption{Case studies of \ours, including WSI Summarized report, Gene Summarized report, Final Result, and Survival Time gt (Case ID: TCGA-XF-A9SJ).}
    \label{fig:case_study_TCGA-XF-A9SJ}
\end{figure*}

\begin{figure*}[!t]
    \centering
    \includegraphics[width=\textwidth]{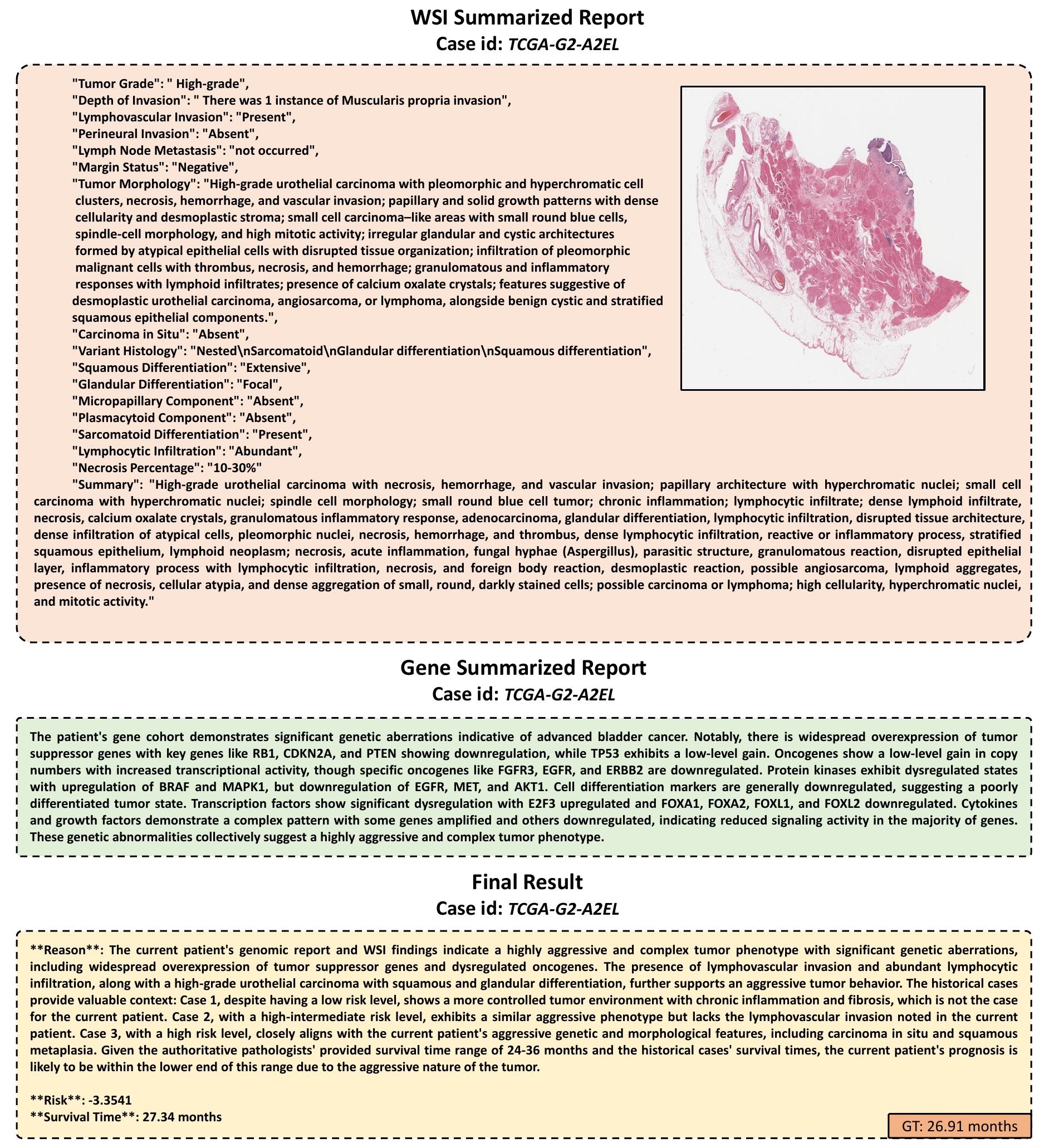}
    \caption{Case studies of \ours, including WSI Summarized report, Gene Summarized report, Final Result, and Survival Time gt (Case ID: TCGA-G2-A2EL).}
    \label{fig:case_study_TCGA-G2-A2EL}
\end{figure*}

To further illustrate the detailed reasoning process of \ours, this section presents several representative case studies. These examples demonstrate how the system analyzes both WSI and genomic data, performs multi-level prognostic reasoning, and integrates morphological cues from WSIs with molecular signatures from gene profiles to generate precise risk predictions. Fig.~\ref{fig:case_study_framework} illustrates the core data involved in WSI analysis, gene analysis, CoT Case Bank construction, and the inference pipeline for the patient with case TCGA-XF-A9SU from the BLCA. Fig.~\ref{fig:case_study_TCGA-XF-A9SJ} and Fig.~\ref{fig:case_study_TCGA-G2-A2EL} focus on the inference stage, presenting SurvAgent’s full WSI reports, gene reports, and the complete CoT-based reasoning process.

\section{Construction of the CoT Case Bank}

\begin{figure*}[t]
    \centering
    \includegraphics[width=\textwidth]{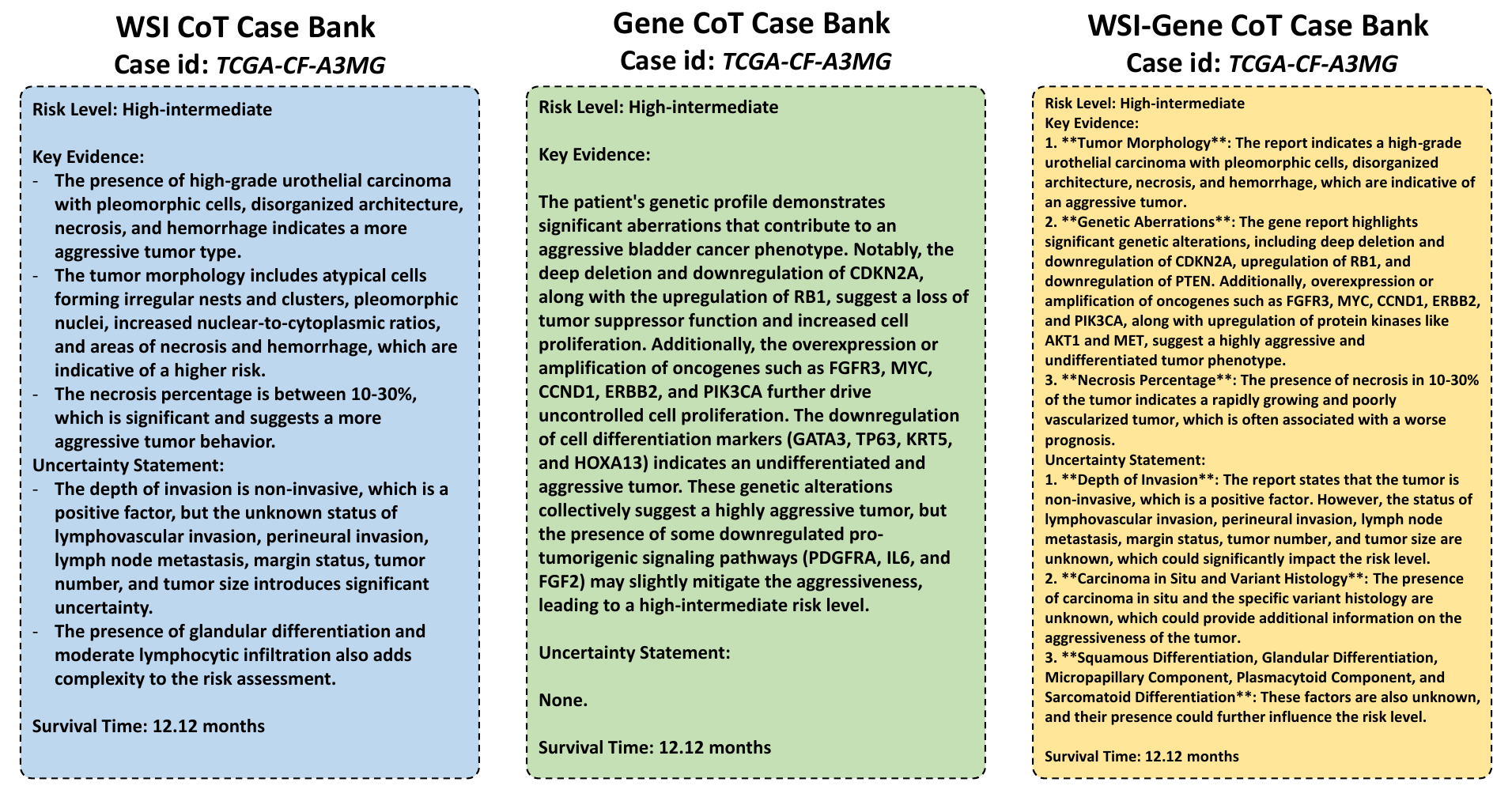}
    \caption{\ours’s CoT Case Bank}
    \label{fig:CoT_case_bank}
\end{figure*}

To enable interpretable reasoning and retrieval-augmented inference within \ours, we construct a unified Chain-of-Thought (CoT) Case Bank. This repository stores structured reasoning trajectories across three complementary levels: WSI-based analysis, gene-level analysis, and integrated WSI–gene reasoning (Fig.~\ref{fig:CoT_case_bank}). Each case follows a standardized schema that includes the assigned risk level, key evidence, and an explicit uncertainty statement, summarizing both the essential prognostic cues and the inherent ambiguity within the reasoning process. The Gene CoT Case Bank captures reasoning grounded in genomic alterations, abnormal expression patterns, and molecular signatures, while the WSI–Gene CoT Case Bank consolidates these perspectives into a coherent, cross-modal prognostic analysis.

\section{Core Prompt Configuration of SurvAgent}

\begin{figure*}[!t]
    \centering
    \includegraphics[width=\textwidth]{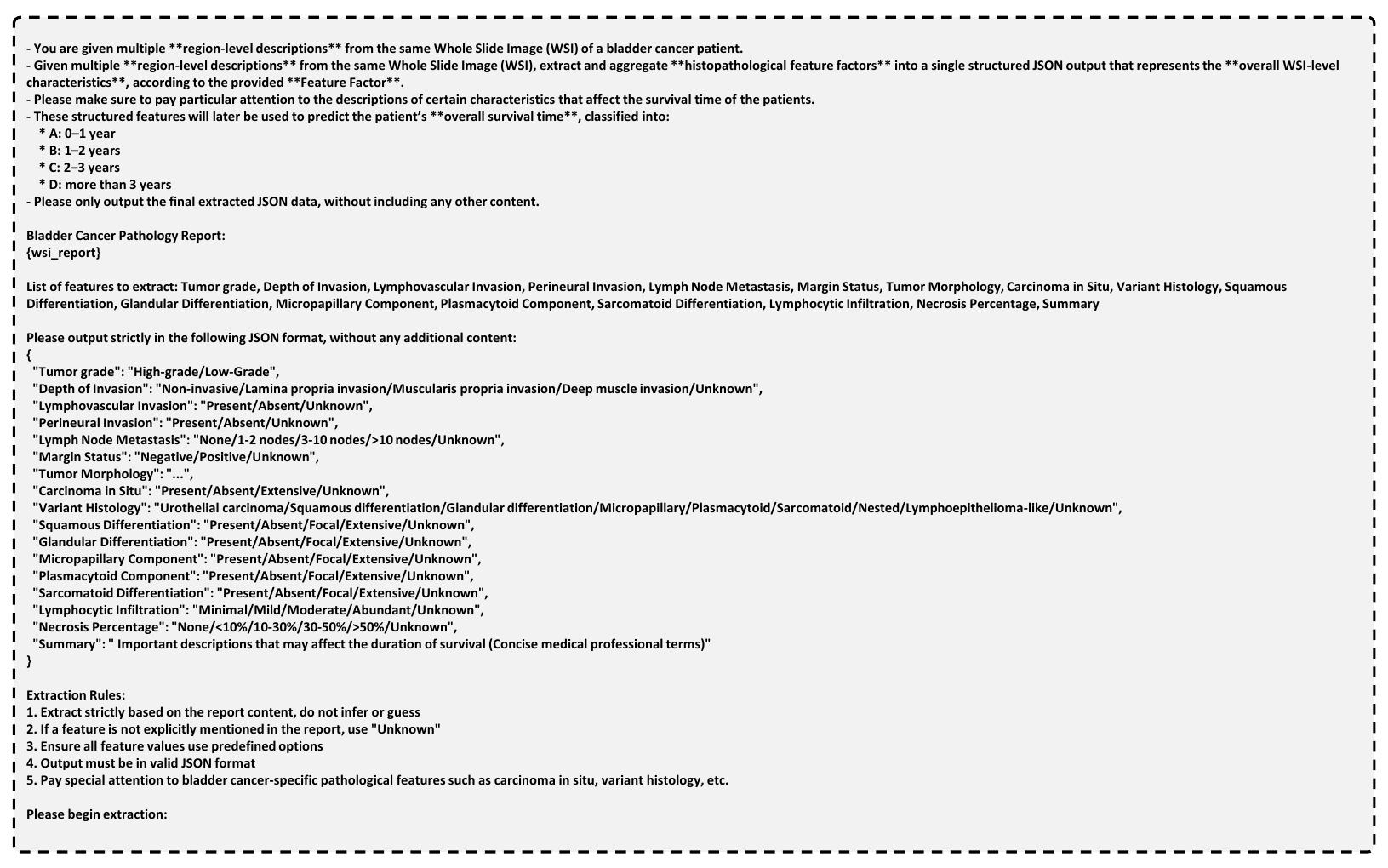}
    \caption{The prompt for extracting a structured WSI report using the WSI attribute checklist.}
    \label{fig:WSI_Attribute_Checklist_Generation_Prompt}
\end{figure*}

\begin{figure*}[!t]
    \centering
    \includegraphics[width=\textwidth]{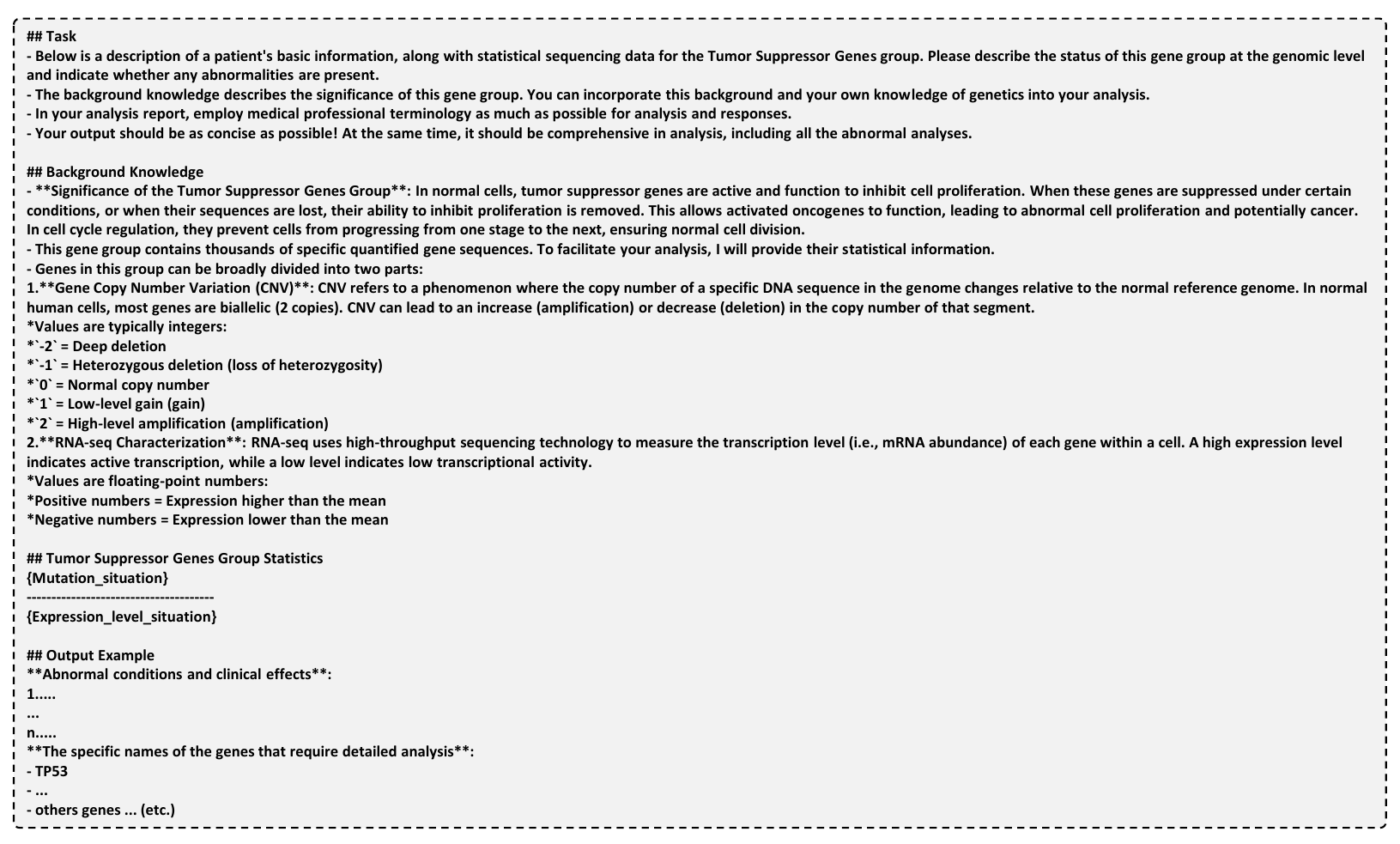}
    \caption{The prompt for statistical feature analysis of six categories of functional genes and key gene selection, using tumor suppressor genes as an example.}
    \label{fig:Gene_Describe_Prompt}
\end{figure*}

\begin{figure*}[!t]
    \centering
    \includegraphics[width=\textwidth]{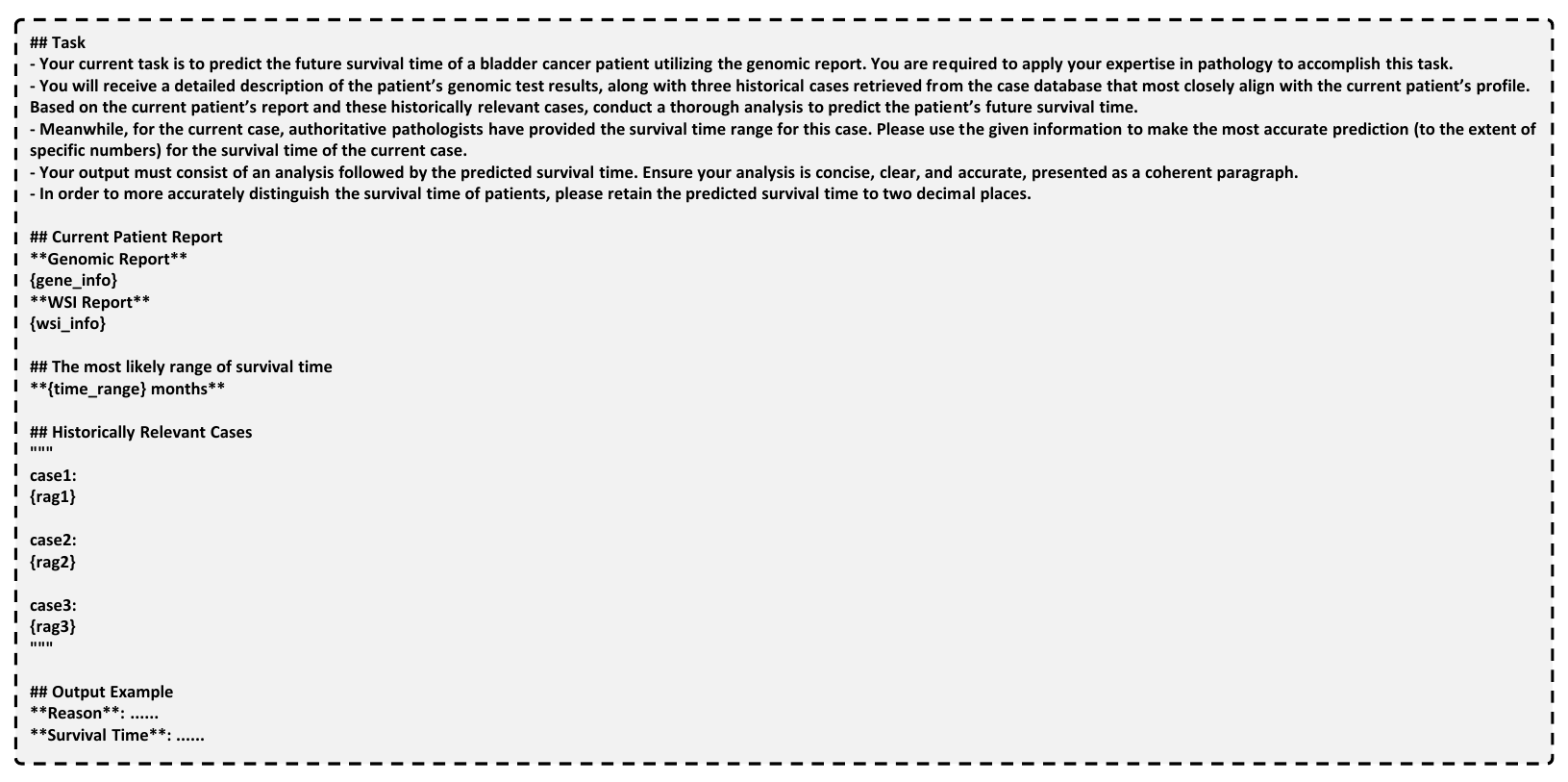}
    \caption{The prompt for the Inference Agent to predict exact survival time within the identified interval in coarse survival intervals by retrieved cases and summarized reports.}
    \label{fig:Dichotomy-Based_Multi-Expert_Agent_Inference_Prompt}
\end{figure*}

\begin{figure*}[!t]
    \centering
    \includegraphics[width=\textwidth]{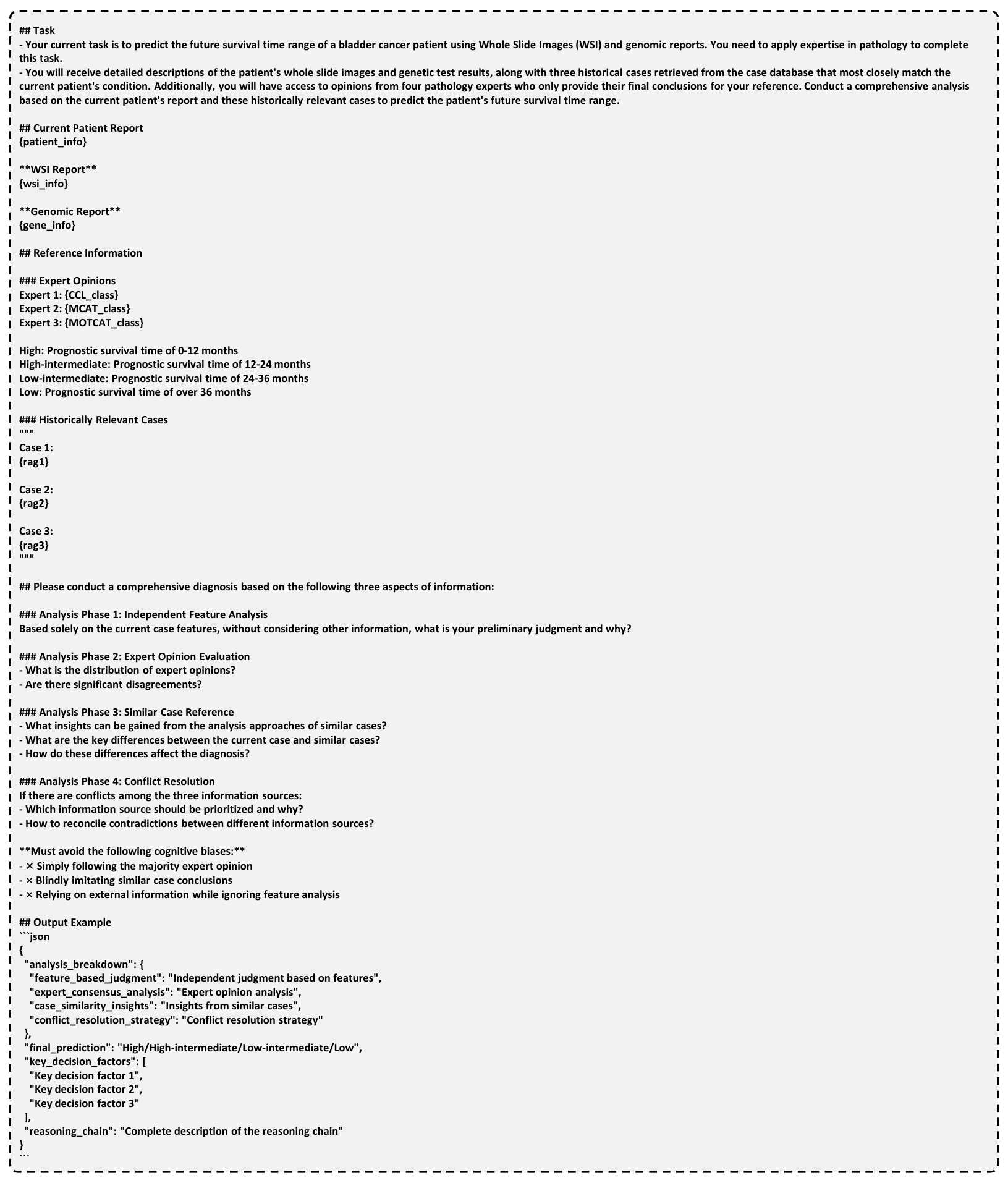}
    \caption{The prompt for the Inference Agent to classify each case into coarse survival intervals by retrieved cases, summarized reports, and predictions from multiple expert survival models.}
    \label{fig:Multi-Expert_Agent_Discussion_Prompt}
\end{figure*}

In this section, we present the core prompt design of \ours. Fig.~\ref{fig:WSI_Attribute_Checklist_Generation_Prompt}--\ref{fig:Multi-Expert_Agent_Discussion_Prompt} illustrate the prompt configurations used in the key stages of \ours. Figure~\ref{fig:WSI_Attribute_Checklist_Generation_Prompt} presents the prompt used by PathAgent to extract a structured WSI report based on the predefined WSI Attribute Checklist. Figure~\ref{fig:Gene_Describe_Prompt} shows the prompt used by GeneAgent to perform statistical feature analysis and key gene selection across six categories of functional genes, with tumor suppressor genes illustrated as an example. Figure~\ref{fig:Dichotomy-Based_Multi-Expert_Agent_Inference_Prompt} displays the prompt used by the Inference Agent to predict the exact survival time within the coarse survival interval determined in the first-stage reasoning, leveraging retrieved similar cases and the current patient’s summarized reports. Figure~\ref{fig:Multi-Expert_Agent_Discussion_Prompt} presents the prompt used by the Inference Agent in the first-stage inference, where it integrates retrieved analogous cases, the patient’s WSI and gene reports, and predictions from multiple expert survival models to determine the coarse survival interval.

\section{Implementation Details}
\subsection{Experimental Setup and Computing Environment}
SurvAgent does not require any additional training, and all results are obtained purely during inference. Experiments are conducted on a computation node equipped with $4 \times$ NVIDIA RTX A6000 GPUs (48 GB each) and an Intel(R) Xeon(R) Gold 6430 CPU. For all expert survival prediction models used in our framework, WSI features are extracted using CLAM~\cite{lu2021data} with the patch level set to 1, and all hyperparameters strictly follow the default settings provided in its open-source implementation. The source code will be released for public access.

\subsection{SurvAgent Architecture}

The SurvAgent framework is entirely developed in-house without relying on any existing agent frameworks. SurvAgent consists of four specialized agents: the Search Agent, PathAgent, GenAgent, and Inference Agent. Their implementations are described below.

Search Agent is built upon DeepSeek-V3.2~\cite{deepseekv32025}, leveraging its web-access capability in combination with a curated pathology knowledge base to generate an initial WSI Attribute Checklist. The resulting checklist is subsequently reviewed and refined by board-certified pathologists.

PathAgent employs PathGen-LLaVA~\cite{sun2024pathgen16m16millionpathology} and Qwen2.5-32B-Instruct~\cite{qwen2025qwen25} as its backbone models. PathGen-LLaVA is responsible for producing expert-level pathological descriptions from WSI image patches, while Qwen2.5-32B-Instruct converts these descriptions into structured pathological attributes, integrates them into a unified report, and performs self-critique on the generated chain-of-thought (CoT) to ensure quality and consistency.

GenAgent is built on Qwen2.5-32B-Instruct, which generates structured gene-level summaries, analyzes statistical properties of functional gene categories, and performs CoT quality verification for gene-related reasoning.

Inference Agent is built on Qwen2.5-32B-Instruct, which is used to perform the final coarse survival interval prediction as well as the precise survival time estimation.

\subsection{Implementation of the Hierarchical WSI CoT-Enhanced Case Bank}

In processing whole-slide images (WSIs), we first apply CLAM at patch level 1 with a patch size of $256 \times 256$ to tile the entire slide, followed by filtering background patches. Patch-level cancer cell detection is then performed using the CHIEF pathology foundation model to identify high-risk regions. Based on the attention scores produced by CHIEF, we apply a DBSCAN clustering procedure to aggregate spatially concentrated high-risk patches, using an epsilon of 4 and a minimum cluster size of 10, thereby determining candidate regions of interest.

WSI examination is conducted in a hierarchical manner across three magnifications: $2.5 \times$ (patch level 3), $10 \times$ (patch level 2), and $20 \times$ (patch level 1). After observing each magnification, PathAgent integrates multi-scale information to form the final WSI-level report used for downstream survival prediction. At $2.5 \times$, PathAgent directly describes the region. At $10 \times$ and $20 \times$, subregions are generated by re-tiling the parent region using $512 \times 512$ windows. Due to the large number of high-resolution patches, we perform CoSMining-based filtering.

For this process, the image-based Self-Path Similarity Matrix is computed via cosine similarity between CHIEF-extracted patch features, while the text-based Self-Path Similarity Matrix is constructed by embedding PathAgent-generated descriptions using the text-embedding-3-large model and computing pairwise cosine similarity. After removing highly redundant patches from both modalities, we take the intersection to obtain the final set of informative subregions.

During the transition from $10 \times$ to $20 \times$, PathAgent autonomously determines whether further magnification is necessary. This behavior is enabled by prompt-based self-reflection, prompting PathAgent to identify uncertainty or ambiguity in its own outputs and decide whether higher-magnification inspection is required.

In the CoT generation stage, PathAgent aligns the multi-scale WSI report with the patient’s ground-truth risk category and survival time to perform reverse reasoning, yielding fine-grained chain-of-thought trajectories that map pathological findings to survival outcomes. To ensure validity, PathAgent further conducts a verification pass in which the CoT is reevaluated without revealing the ground truth, enabling the agent to detect inconsistencies and revise the CoT accordingly.

\subsection{Implementation of the Gene-Stratified CoT-Enhanced Case Bank}

For processing genomic data, we follow prior work to categorize genes into six major functional groups: Tumor Suppressor Genes, Oncogenes, Protein Kinases, Cell Differentiation Markers, Transcription Factors, and Cytokines and Growth Factors. Owing to the large number of genes and genomic fragments, we first conduct statistical quantification to capture the global expression patterns of each gene category.

Specifically, genomic information is divided into DNA-level structural variation data (CNV), RNA-level expression data (RNA-seq), and other special genomic fragments. For RNA-seq expression profiles, the mean and median expression values of each gene are computed to characterize the overall expression distribution of the corresponding gene class. For CNV data, we quantify the mutation rate—defined as the proportion of samples exhibiting mutations such as point mutations, insertions/deletions, or amplifications/deletions—to assess the overall structural variability within each gene category.

After statistical quantification, GenAgent performs a high-level analysis of each gene class to understand its global expression characteristics. GenAgent then generates a preliminary class-level gene expression report and identifies specific genes that require additional inspection. For these selected genes, their raw expression values are retrieved, and GenAgent accesses biological function information via the Gene Knowledge Base constructed using the python library mygene*. By integrating the statistical summaries with functional gene annotations, GenAgent performs a systematic, coarse-to-fine analysis of expression behaviors, ultimately producing a detailed report for each gene category.

The reports from all categories are then consolidated into a unified genomic feature report, which serves as input for downstream inference. The process for generating chain-of-thought (CoT) explanations follows the same procedure described in the previous subsection.

\subsection{Implementation of the Dichotomy-Based Multi-Expert Agent Inference Module}

The inference process of SurvAgent is conducted in two stages: (1) estimating a coarse survival interval, and (2) predicting the exact survival time. In the first stage, the reasoning of the Inference Agent is enhanced by both retrieval-augmented generation (RAG) and the outputs of multiple expert survival prediction models. In the second stage, for fine-grained survival time prediction, only RAG-based retrieval information is used to augment the inference process.

After obtaining the patient’s integrated WSI report and genomic report, we perform cosine-similarity–based retrieval over the previously constructed WSI–Gene CoT Case Bank to identify the top three most similar historical cases. Each retrieved case provides its WSI report, gene report, and corresponding chain-of-thought (CoT) annotations.

Following the definition of risk scores from expert survival models, we map each model’s predicted risk value into one of four risk strata based on quartiles. These strata align with the target survival intervals: High (0–12 months), High-intermediate (12–24 months), Low-intermediate (24–36 months), and Low (36+ months). The Inference Agent integrates the current patient’s multimodal information, retrieved case evidence, and risk assessments from multiple expert models to determine the patient’s final risk stratum.

Once the coarse survival interval is established, we further retrieve the actual survival times of similar cases. These values are provided as additional prompts to the Inference Agent to facilitate precise survival time prediction at a monthly resolution.

\end{document}